%% file: main.tex
\documentclass[journal,10pt,twocolumn]{IEEETran}
\usepackage[utf8]{inputenc}

\usepackage{threeparttable}
\usepackage{cite}
\usepackage{amsmath,amssymb,amsfonts}
\usepackage{algorithmic}
\usepackage{graphicx}
\usepackage{textcomp}
\usepackage{indentfirst}
\usepackage{graphicx}
\usepackage{csquotes}
\usepackage{multirow}
\usepackage{tabularx}
\usepackage{float}
\usepackage{color}
\usepackage{framed}
\usepackage{comment}
\usepackage{caption2}
\usepackage[boxed]{algorithm2e}
\usepackage{bbding}

\renewcommand{\figurename}{Figure}
\def\tablename{Table}

\begin{document}

\title{Comparing SNNs and RNNs on Neuromorphic Vision Datasets: Similarities and Differences}

\author{Weihua He$^{a,b}$, YuJie Wu$^{a,c}$, Lei Deng$^{b,*}$, Guoqi Li$^{a,c}$, Haoyu Wang$^{a}$, Yang Tian$^{d}$, \\Wei Ding$^{a}$, Wenhui Wang$^{a}$, Yuan Xie$^{b}$\\ \vspace{10pt}
$^a$Department of Precision Instrument, Tsinghua University, Beijing 100084, China\\
$^b$Department of Electrical and Computer Engineering, University of California, Santa Barbara, CA 93106, USA\\
$^c$Center for Brain Inspired Computing Research, Tsinghua University, Beijing 100084, China\\
$^d$Lab of Cognitive Neuroscience, THBI, Tsinghua University, Beijing 100084, China
\thanks{Email addresses: hewh16@mails.tsinghua.edu.cn (W. He), wu-yj16@mails.tsinghua.edu.cn (Y. Wu), leideng@ucsb.edu (L. Deng), liguoqi@mail.tsinghua.edu.cn (G. Li), haoyu-wa16@mails.tsinghua.edu.cn (H. Wang), tianyang16@mails.tsinghua.edu.cn (Y. Tian), dingw17@mails.tsinghua.edu.cn (W. Ding), wwh@mail.tsinghua.edu.cn (W. Wang), yuanxie@ucsb.edu (Y. Xie). $^{*}$Corresponding author: Lei Deng.}
}

\maketitle

\begin{abstract}
Neuromorphic data, recording frameless spike events, have attracted considerable attention for the spatiotemporal information components and the event-driven processing fashion. Spiking neural networks (SNNs) represent a family of event-driven models with spatiotemporal dynamics for neuromorphic computing, which are widely benchmarked on neuromorphic data. Interestingly, researchers in the machine learning community can argue that recurrent (artificial) neural networks (RNNs) also have the capability to extract spatiotemporal features although they are not event-driven. Thus, the question of ``what will happen if we benchmark these two kinds of models together on neuromorphic data'' comes out but remains unclear.

In this work, we make a systematic study to compare SNNs and RNNs on neuromorphic data, taking the vision datasets as a case study. First, we identify the similarities and differences between SNNs and RNNs (including the vanilla RNNs and LSTM) from the modeling and learning perspectives. To improve comparability and fairness, we unify the supervised learning algorithm based on backpropagation through time (BPTT), the loss function exploiting the outputs at all timesteps, the network structure with stacked fully-connected or convolutional layers, and the hyper-parameters during training. Especially, given the mainstream loss function used in RNNs, we modify it inspired by the rate coding scheme to approach that of SNNs. Furthermore, we tune the temporal resolution of datasets to test model robustness and generalization. At last, a series of contrast experiments are conducted on two types of neuromorphic datasets: DVS-converted (N-MNIST) and DVS-captured (DVS Gesture). Extensive insights regarding recognition accuracy, feature extraction, temporal resolution and contrast, learning generalization, computational complexity and parameter volume are provided, which are beneficial for the model selection on different workloads and even for the invention of novel neural models in the future.
\end{abstract}


\input{text/Introduction.tex}
\input{text/Preliminary.tex}
\input{text/Methodology.tex}
\input{text/Results.tex}
\input{text/Conclusion.tex}

\section*{Acknowledgment}
The work was partially supported by National Science Foundation (Grant No. 1725447), Beijing Academy of Artificial Intelligence (BAAI), Tsinghua University Initiative Scientific Research Program, and a grant from the Institute for Guo Qiang, Tsinghua University.

\bibliography{main}

\end{document}

%% file: text/Introduction.tex
\section{Introduction}\label{sec:introduction}

Neuromorphic vision datasets \cite{orchard2015converting,barranco2016dataset,li2017cifar10} sense the dynamic change of pixel intensity and record the resulting spike events using dynamic vision sensors (DVS) \cite{delbruck2008frame,lichtsteiner2008128,serrano2013128,yang2015dynamic}. Compared to conventional frame-based vision datasets, the frameless neuromorphic vision datasets have rich spatiotemporal components by interacting the spatial and temporal information and follow the event-driven processing fashion triggered by binary spikes. Owing to these unique features, neuromorphic data have attracted considerable attention in many applications such as visual recognition \cite{orchard2015hfirst,zhao2014feedforward,ramesh2019dart,xiao2019event}\cite{wu2018spatio,kaiser2018synaptic,amir2017low},  motion segmentation \cite{mishra2017saccade}, tracking control \cite{drazen2011toward,conradt2009embedded,hu2016dvs}, robotics \cite{delbruck2013robotic}, etc. Currently, there are two types of neuromorphic vision datasets: one is converted from static datasets by scanning each image in front of DVS cameras, e.g. N-MNIST \cite{orchard2015converting} and CIFAR10-DVS \cite{li2017cifar10}; the other is directly captured by DVS cameras from moving objects, e.g. DVS Gesture \cite{amir2017low}.

Spiking neural networks (SNNs) \cite{maass1997networks}, inspired by brain circuits, represent a family of models for neuromorphic computing. Each neuron in an SNN model updates the membrane potential based on its memorized state and current inputs, and fires a spike when the membrane potential crosses a threshold. The spiking neurons communicate with each other using binary spike events rather than continuous activations in artificial neural networks (ANNs), and an SNN model carries both spatial and temporal information. The rich spatiotemporal dynamics and event-driven paradigm of SNNs hold great potential in efficiently handling complex tasks such as spike pattern recognition \cite{orchard2015hfirst,wu2019direct}, optical flow estimation \cite{haessig2018spiking}, and simultaneous localization and map (SLAM) building \cite{vidal2018ultimate}, which motivates their wide deployment on low-power neuromorphic devices \cite{merolla2014million,davies2018loihi,pei2019towards}. Since the behaviors of SNNs naturally match the characteristics of neuromorphic data, a considerable amount of literature benchmark the performance of SNNs on neuromorphic datasets \cite{wu2018spatio,shrestha2018slayer,jin2018hybrid,deng2020rethinking}.

Originally, neuromorphic computing and machine learning are two domains developing in parallel and are usually independent of each other. It seems that this situation is changing as more interdisciplinary researches emerge \cite{pei2019towards,deng2020tianjic,deng2020rethinking}. In this context, researchers in the machine learning community can argue that SNNs are not unique for the processing of neuromorphic data. The reason is that recurrent (artificial) neural networks (RNNs) can also memorize previous states and behave spatiotemporal dynamics, even though they are not event-driven. By treating the spike events as normal binary values in $\{0,~1\}$, RNNs are able to process neuromorphic datasets too. In essence, RNNs have been widely applied in many tasks with timing sequences such as language modeling \cite{mikolov2012statistical}, speech recognition \cite{miao2015eesen}, and machine translation \cite{cho2014learning}; whereas, there are rare studies that evaluate the performance of RNNs on neuromorphic data, thus the mentioned debate still remains open. 

In this work, we try to answer what will happen when benchmarking SNNs and RNNs together on neuromorphic data, taking the vision datasets as a case study. First, we identify the similarities and differences between SNNs and RNNs from the modeling and learning perspectives. For comparability and fairness, we unify several things: i) supervised learning algorithm based on backpropagation through time (BPTT); ii) loss function inspired by the SNN-oriented rate coding scheme; iii) network structure based on stacked fully-connected (FC) or convolutional (Conv) layers; iv) hyper-parameters during training. Moreover, we tune the temporal resolution of neuromorphic vision datasets to test the model robustness and generalization. At last, we conduct a series of contrast experiments on typical neuromorphic vision datasets and provide extensive insights. Our work holds potential in guiding the model selection on different workloads and stimulating the invention of novel neural models. For clarity, we summarize our contributions as follows:

\begin{itemize}
\item We present the first work that systematically compares SNNs and RNNs on neuromorphic datasets.

\item We identify the similarities and differences between SNNs and RNNs, and unify the learning algorithm, loss function, network structure, and training hyper-parameters to ensure the comparability and fairness. Especially, we modify the mainstream loss function of RNNs to approach that of SNNs and tune the temporal resolution of neuromorphic vision datasets to test model robustness and generalization.

\item On two kinds of typical neuromorphic vision datasets: DVS-converted (N-MNIST) and DVS-captured (DVS Gesture), we conduct a series of contrast experiments that yield extensive insights regarding recognition accuracy, feature extraction, temporal resolution and contrast, learning generalization, computational complexity and parameter volume (detailed in Section \ref{sec:result} and summarized in Section \ref{sec:conclusion}), which are beneficial for future model selection and construction.
\end{itemize}

The rest of this paper is organized as follows: Section \ref{sec:preliminary} introduces some preliminaries of neuromorphic vision datasets, SNNs, and RNNs; Section \ref{sec:methodology} details our methodology to make SNNs and RNNs comparable and ensure the fairness; Section \ref{sec:result} shows the experimental results and provides our insights; Finally, Section \ref{sec:conclusion} concludes and discusses the paper.

%% file: text/Preliminary.tex
\section{Preliminaries}\label{sec:preliminary}

\subsection{Neuromorphic Vision Datasets}\label{sec:preliminary:dataset}

A neuromorphic vision dataset consists of many spike events, which are triggered by the intensity change (increase or decrease) of each pixel in the sensing field of the DVS camera \cite{delbruck2008frame,lichtsteiner2008128,hu2016dvs}. A DVS camera records the spike events in two channels according to the different change directions, e.g. the On channel for intensity increase and the Off channel for intensity decrease. The whole spike train in a neuromorphic vision dataset can be represented as an $H \times W \times 2 \times T_0$ sized spike pattern, where $H,~W$ are the height and width of the sensing field, respectively, $T_0$ stands for the length of recording time, and ``2'' indicates the two channels. As mentioned in Introduction, currently there are two types of neuromorphic vision datasets: DVS-converted and DVS-captured, which are detailed as below.

\begin{figure}[!htbp]
\centering     
\includegraphics[width=0.4\textwidth]{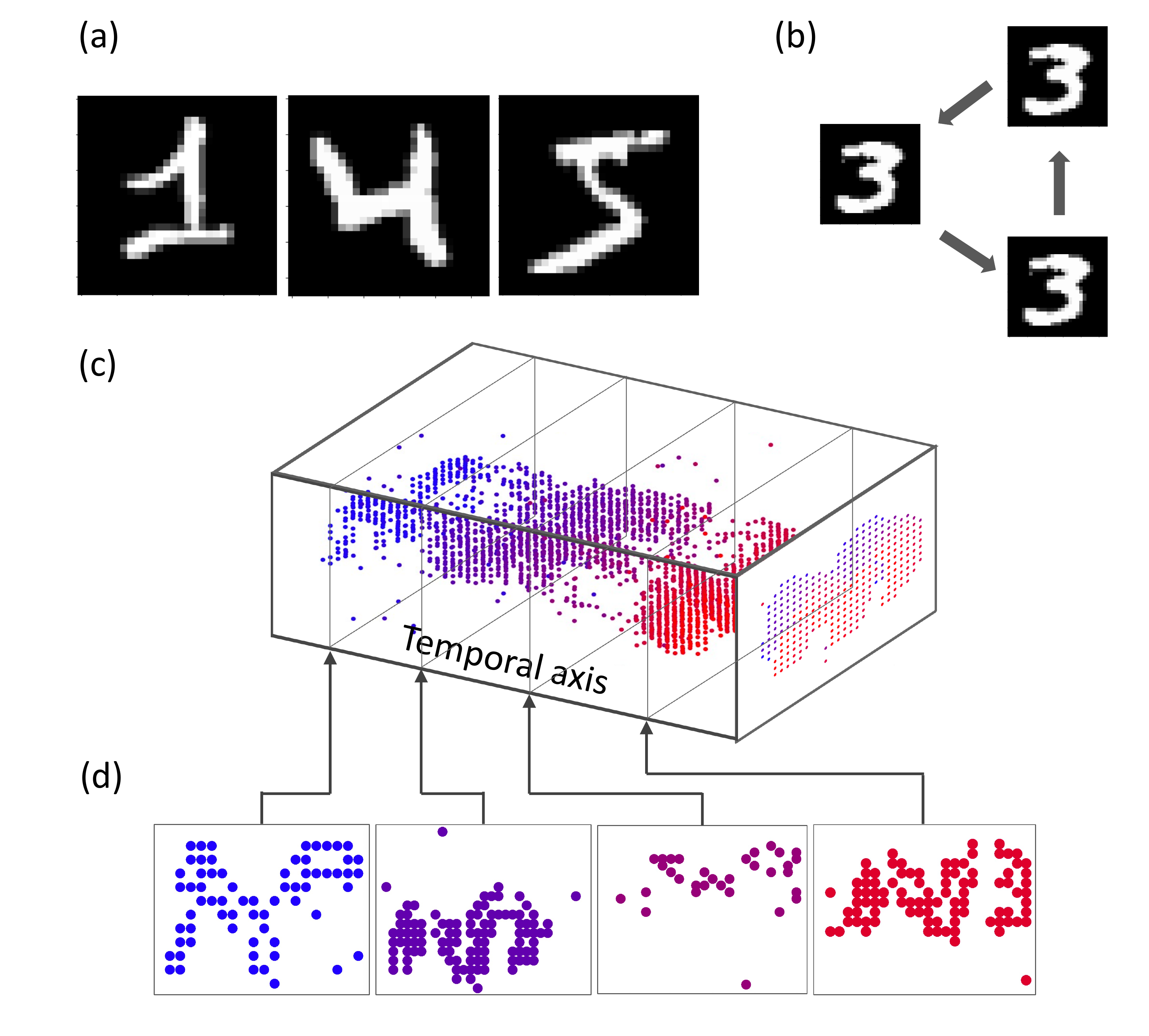}
\caption{\textbf{An example of DVS-converted dataset named N-MNIST:} (a) original static images in the MNIST dataset; (b) the repeated closed-loop smooth (RCLS) movement of static images; (c) the resulting spike pattern recorded by DVS; (d) slices of spike events at different timesteps. The blue and red colors denote the On and Off channels, respectively.} \label{fig:NMNIST} 
\end{figure}

\textbf{DVS-Converted Dataset.} Generally, DVS-converted datasets are converted from frame-based static image datasets. The spike events in a DVS-converted dataset are acquired by scanning each image with repeated closed-loop smooth (RCLS) movement in front of a DVS camera \cite{serrano2015poker,li2017cifar10}, where the movement incurs pixel intensity changes. Figure \ref{fig:NMNIST} illustrates a DVS-converted dataset named N-MNIST \cite{orchard2015converting}. The original MNIST dataset includes 60000 $28\times28$ static images of handwritten grayscale digits for training and extra 10000 for testing; accordingly, the DVS camera converts each image in MNIST into a $34 \times 34 \times 2\times T_0$ spike pattern in N-MNIST. The larger sensing field is caused by the RCLS movement. Compared to the original frame-based static image dataset, the converted frameless dataset contains additional temporal information while retaining the similar spatial information. Nevertheless, the extra temporal information cannot become dominant due to the static information source, and some works even point out that the DVS-converted datasets might be not good enough to benchmark SNNs \cite{iyer2018neuromorphic,deng2020rethinking}. 

\begin{figure}[!htbp]
\centering     
\includegraphics[width=0.4\textwidth]{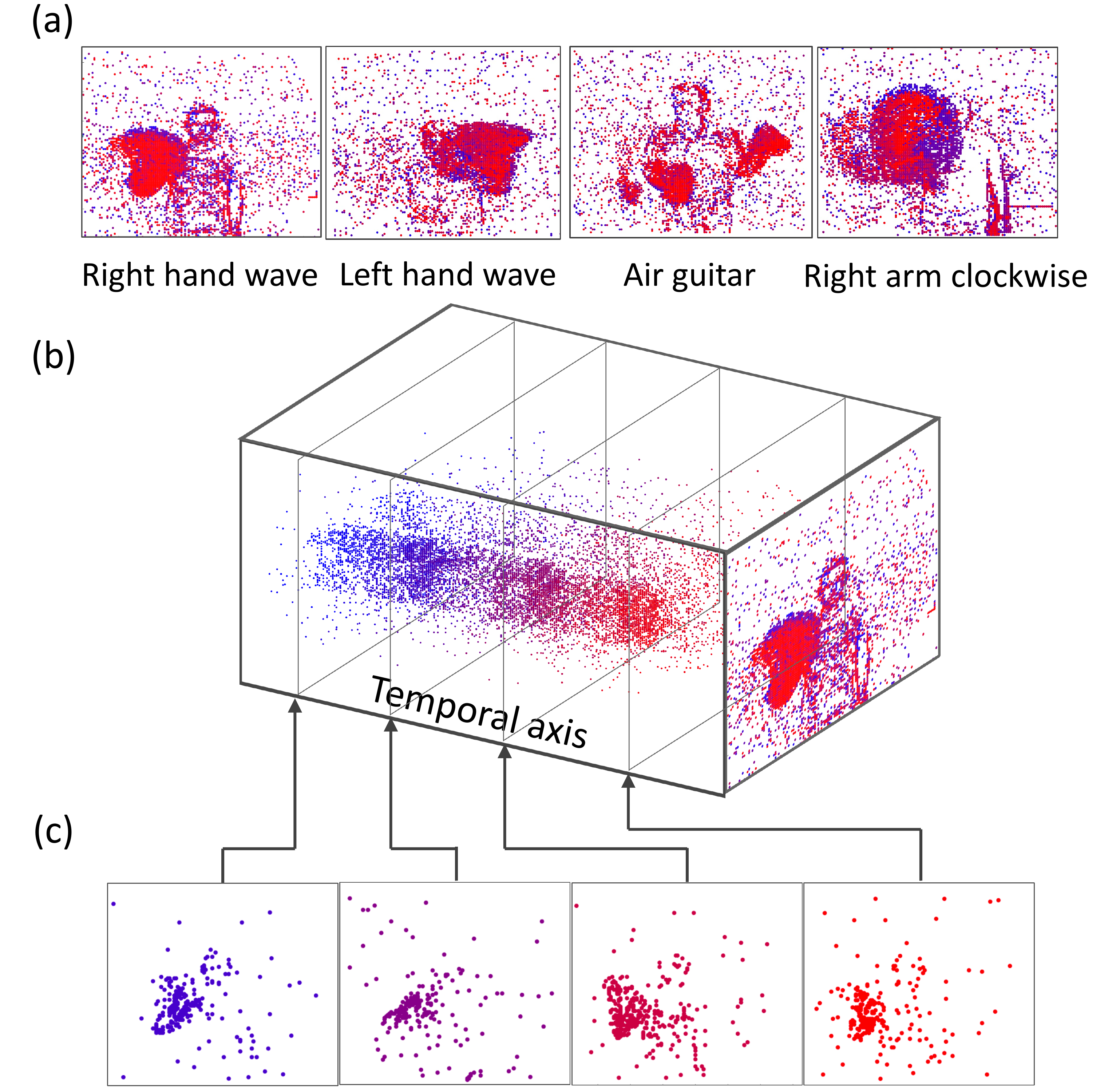}
\caption{\textbf{An example of DVS-captured dataset named DVS Gesture:} (a) spike pattern examples, where all spike events within a time window $t_{w}$ are compressed into a single static image for visualization; (b) the spike pattern recorded by DVS; (c) slices of spike events at different timesteps, where the right hand lies at a different location in each slice thus forming a hand wave along the time dimension. The blue and red colors denote the On and Off channels, respectively.} \label{fig:DVSGesture} 
\end{figure}

\textbf{DVS-Captured Dataset.} In contrast, DVS-captured datasets generate spike events via natural motion rather than the simulated movement used in the generation of DVS-converted datasets. Figure \ref{fig:DVSGesture} depicts a DVS-captured dataset named DVS Gesture \cite{amir2017low}. There are 11 hand and arm gestures performed by one subject in each trail, and there are total 122 trails in the dataset. Three lighting conditions including natural light, fluorescent light, and LED light are selected to control the effects of shadow and flicker on the DVS camera, providing a bias improvement for the data distribution. Different from the DVS-converted datasets, both temporal and spatial information in DVS-captured datasets are featured as essential components.

\subsection{Spiking Neural Networks}

There are several different spiking neuron models such as leaky integrate and fire (LIF) \cite{abbott1999lapicque}, Izhikevich \cite{izhikevich2003simple}, and Hodgkin-Huxley \cite{hodgkin1952quantitative}, among which LIF is the most widely used in practice due to the lower complexity \cite{izhikevich2004model}. In this work, we adopt the mainstream solution, i.e. taking LIF for neuron simulation. By connecting many spiking neurons through synapses, we can construct an SNN model. In this paper, for simplicity, we just consider feedforward SNNs with stacked FC or Conv layers.

There are two state variables in a LIF neuron: membrane potential ($u$) and output activity ($o$). $u$ is a continuous value while $o$ can only be a binary value, i.e. firing a spike or not. The behaviors of an SNN layer can be described as
\begin{equation}
    \label{equ:SNN_layer_1}
    \begin{cases}
\tau \frac{d\pmb{u}^{n}(t)}{dt} = -\pmb{u}^{n}(t) + \pmb{W}^{n}\pmb{o}^{n-1}(t)\\
\begin{cases}
o_i^{n}(t)=1~\&~u_i^{n}(t)=u_0,~\text{if}~u_i^{n}(t)\geq u_{th}\\
o_i^{n}(t)=0,~\text{if}~u_i^{n}(t)< u_{th}~
\end{cases}
\end{cases}
\end{equation}
where $t$ denotes time, $n$ and $i$ are indices of layer and neuron, respectively, $\tau$ is a time constant, and $\pmb{W}$ is the synaptic weight matrix between two adjacent layers. The neuron fires a spike and resets $u=u_0$ only when $u$ exceeds a firing threshold ($u_{th}$), otherwise, the membrane potential would just leak. Notice that $\pmb{o}^{0}(t)$ denotes the network input.

\subsection{Recurrent Neural Networks}

In this work, RNNs mainly mean recurrent ANNs rather than SNNs. We select two kinds of RNN models in this work: one is the vanilla RNN and the other is the modern RNN named long short-term memory (LSTM).

\textbf{Vanilla RNN.} RNNs introduce temporal dynamics via recurrent connections. There is only one continuous state variable in a vanilla RNN neuron, called hidden state ($h$). The behaviors of a vanilla RNN layer follow
\begin{equation}
\label{equ:vanilla_RNN_layer}
\pmb{h}^{t,n}=\theta(\pmb{W}_1^{n}\pmb{h}^{t,n-1} + \pmb{W}_2^{n}\pmb{h}^{t-1,n} + \pmb{b}^{n})
\end{equation}
where $t$ and $n$ denote the indices of timestep and layer, respectively, $\pmb{W}_1$ is the weight matrix between two adjacent layers, $\pmb{W}_2$ is the intra-layer recurrent weight matrix, and $\pmb{b}$ is a bias vector. $\theta(\cdot)$ is an activation function, which can be the $tanh(\cdot)$ function in general for vanilla RNNs. Similar to the $\pmb{o}^{0}(t)$ for SNNs, $\pmb{h}^{t,0}$ also denotes the network input of RNNs, i.e. $\pmb{x}^t$.

\textbf{Long Short-Term Memory (LSTM).} LSTM is designed to improve the long-term temporal dependence over vanilla RNNs by introducing complex gates to alleviate gradient vanishing \cite{hochreiter1997long,gers1999learning}. An LSTM layer can be described as
\begin{equation}
\label{equ:LSTM_layer}
\begin{cases}
\pmb{f}^{t,n}=\sigma_f(\pmb{W}_{f1}^n\pmb{h}^{t,n-1}+\pmb{W}_{f2}^n\pmb{h}^{t-1,n} + \pmb{b}_f^n) \\
\pmb{i}^{t,n}=\sigma_i(\pmb{W}_{i1}^n\pmb{h}^{t,n-1}+\pmb{W}_{i2}^n\pmb{h}^{t-1,n} + \pmb{b}_i^n) \\
\pmb{o}^{t,n}=\sigma_o(\pmb{W}_{o1}^n\pmb{h}^{t,n-1}+\pmb{W}_{o2}^n\pmb{h}^{t-1,n} + \pmb{b}_o^n) \\
\pmb{g}^{t,n}=\theta_g(\pmb{W}_{g1}^n\pmb{h}^{t,n-1}+\pmb{W}_{g2}^n\pmb{h}^{t-1,n} + \pmb{b}_g^n) \\
\pmb{c}^{t,n}=\pmb{c}^{t-1,n}\circ \pmb{f}^{t,n}+\pmb{g}^{t,n}\circ \pmb{i}^{t,n} \\
\pmb{h}^{t,n}=\theta_c(\pmb{c}^{t,n})\circ \pmb{o}^{t,n}
\end{cases}
\end{equation}
where $t$ and $n$ denote the indices of timestep and layer, respectively, $\pmb{f}$, $\pmb{i}$, $\pmb{o}$ are the states of forget, input, and output gates, respectively, and $\pmb{g}$ is the input activation. Each gate has its own weight matrices and bias vector. $\pmb{c}$ and $\pmb{h}$ are cellular and hidden states, respectively. $\sigma(\cdot)$ and $\theta(\cdot)$ are $sigmoid$ and $tanh$ functions, respectively, and $\circ$ is the Hadamard product.

%% file: text/Methodology.tex
\section{Methodology}\label{sec:methodology}

To avoid ambiguity, we would like to emphasize again that our ``SNNs vs. RNNs'' in this work is defined as ``feedforward SNNs vs. recurrent ANNs''. For simplicity, we only select two representatives from the RNN family, i.e. vanilla RNNs and LSTM. In this section, we first rethink the similarities and differences between SNNs and RNNs from the modeling and learning perspectives, and discuss how to ensure the comparability and fairness.

\begin{figure*}[!htbp]
\centering     
\includegraphics[width=0.88\textwidth]{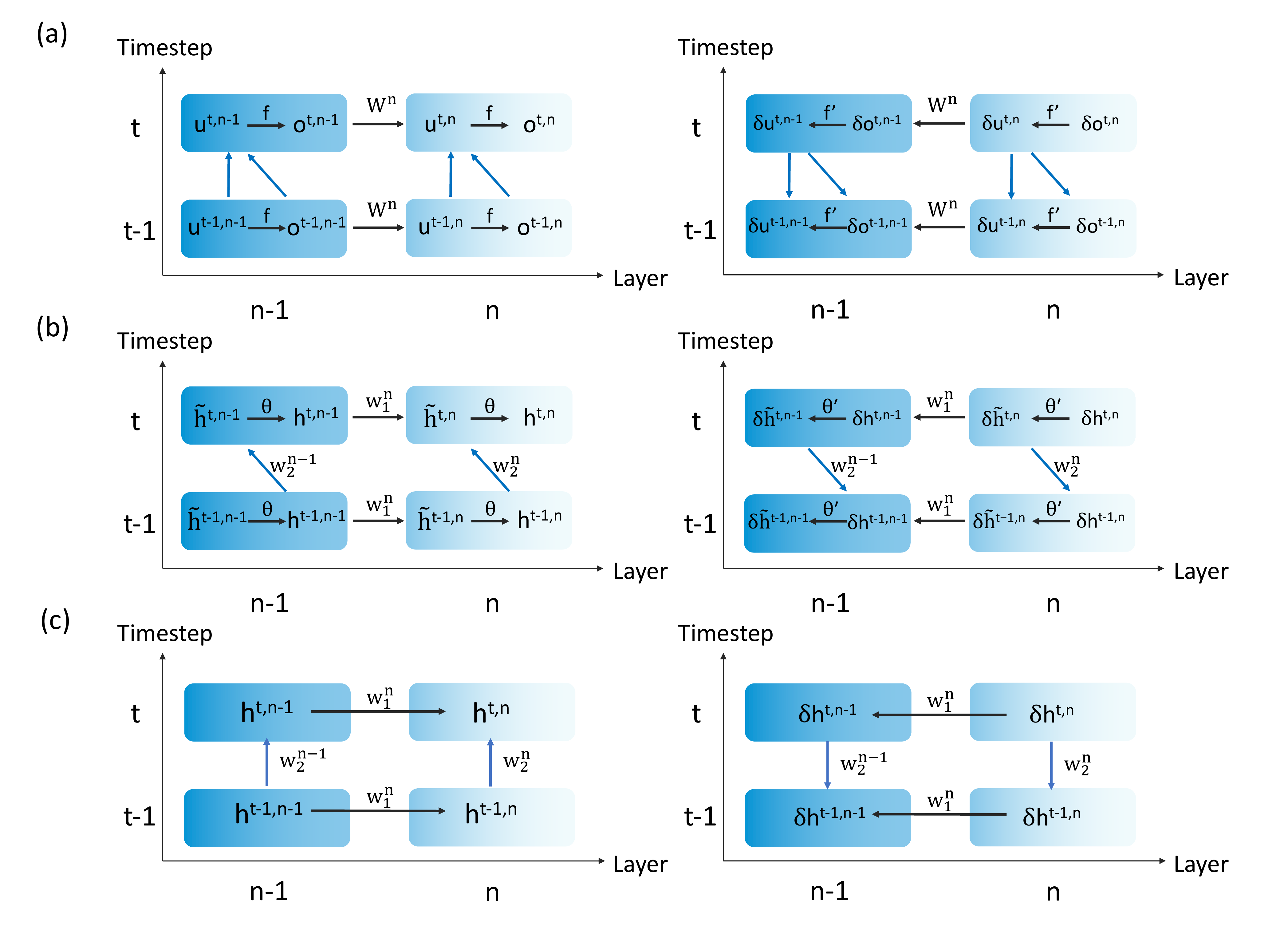}
\vspace{-10pt}
\caption{\textbf{Information propagation paths of (a) SNNs, (b) vanilla RNNs, and (c) LSTM in both forward and backward passes.} For clarity, we do not show the dataflow inside LSTM neurons.}
\label{fig:dataflow} 
\end{figure*}

\subsection{Rethinking the Similarities}

Before analysis, we first convert Equation (\ref{equ:SNN_layer_1}) to its iterative version to make it compatible with the format in Equation (\ref{equ:vanilla_RNN_layer})-(\ref{equ:LSTM_layer}). This can be achieved by solving the first-order differential equation in Equation (\ref{equ:SNN_layer_1}), which yields
\begin{equation}
    \label{equ:SNN_layer_2}
    \begin{cases}
    \pmb{u}^{t,n}= e^{-\frac{dt}{\tau}}\pmb{u}^{t-1,n}\circ(1-\pmb{o}^{t-1,n}) + \pmb{W}^{n}\pmb{o}^{t,n-1} \\
    \pmb{o}^{t,n}= f(\pmb{u}^{t,n}-u_{th})
    \end{cases}
\end{equation}
where $e^{-\frac{dt}{\tau}}$ reflects the leakage effect of the membrane potential, and $f(\cdot)$ is a step function that satisfies $f(x)=1$ when $x\geq 0$, otherwise $f(x)=0$. This iterative LIF model incorporates all behaviors of a spiking neuron, including integration, fire, and reset. 

Now, from Equation (\ref{equ:vanilla_RNN_layer})-(\ref{equ:SNN_layer_2}), it can be seen that the models of SNNs and RNNs are quite similar, involving both temporal and spatial dimensions. Figure \ref{fig:dataflow} further visualizes the information propagation paths of SNNs, vanilla RNNs, and LSTM in both forward and backward passes. Here we denote the hidden state before $\theta$ in vanilla RNNs as $\tilde{h}$ and denote the gate states before $\sigma,~\theta$ in LSTM as $\tilde{f},~\tilde{i},~\tilde{o},~\tilde{g}$. 

\textbf{Spatiotemporal Dynamics in the Forward Pass.} First, the forward propagation paths of SNNs and vanilla RNNs are similar if $u$ and $o$ of SNNs are regarded as $\tilde{h}$ and $h$ of vanilla RNNs, respectively. Second, for LSTM, there are more intermediate states inside a neuron, including $\tilde{f},~\tilde{i},~\tilde{o},~\tilde{g}$ and the cellular state. Although the neuron becomes complicated, the overall spatiotemporal path is still similar if we just pay attention to the propagation of the hidden state $h$. Interestingly, the internal membrane potential of each spiking neuron can directly affect the neuronal state at the next timestep, which differs them from vanilla RNNs but acts similarly as the forget gate of LSTM.

\textbf{Spatiotemporal Backpropagation.} For SNNs, the learning algorithms significantly vary in literature, for example, including unsupervised learning \cite{diehl2015unsupervised}, ANN-to-SNN conversion \cite{diehl2015fast}, and supervised learning \cite{lee2016training,wu2018spatio,wu2019direct}. Since RNNs are usually trained by the gradient-descent-based supervised algorithm in the machine learning domain, we select a recent BPTT-inspired spatiotemporal backpropagation algorithm \cite{wu2018spatio,wu2019direct} for SNNs to make our comparison fair. 

Also from Figure \ref{fig:dataflow}, it can be seen that the gradient propagation paths of SNNs, vanilla RNNs, and LSTM also follow the similar spatiotemporal fashion. Moreover, we detail the backpropagation formula of each model for better understanding. Notice that the variable $\delta$ denotes the gradient, for example, $\delta o=\frac{\partial L}{\partial o}$ where $L$ is the loss function of the network. For SNNs, we have
\begin{equation}
\label{equ:SNN_bp}
\begin{cases}
\delta \pmb{o}^{t,n} = (\pmb{W}^{n+1})^T\delta \pmb{u}^{t,n+1} - e^{-\frac{dt}{\tau}} \delta \pmb{u}^{t+1,n}\circ \pmb{u}^{t,n}
\\
\delta \pmb{u}^{t,n} = \delta \pmb{o}^{t,n}\circ f' + e^{-\frac{dt}{\tau}} \delta \pmb{u}^{t+1,n}\circ (1-\pmb{o}^{t,n})
\end{cases}
\end{equation}
where the firing function is non-differentiable. To this end, a Dirac-like function is introduced to approximate its derivative \cite{wu2018spatio}. Specifically, $f'(\cdot)$ can be calculated by
\begin{equation}
    \label{equ:fire_bp}
    f'(u-u_{th}) \approx
    \begin{cases}
    \frac{1}{a},~~|u-u_{th}|\leq \frac{a}{2} \\
    0,~~otherwise
    \end{cases}
\end{equation}
where $a$ is a hyper-parameter to control the gradient width when passing the firing function during backpropagation. For vanilla RNNs, we have a similar format as follows
\begin{equation}
\label{equ:vanilla_RNN_bp}
\delta \pmb{h}^{t,n}=(\pmb{W}_1^{n+1})^T(\delta \pmb{h}^{t,n+1}\circ \theta') + (\pmb{W}_2^{n})^T(\delta \pmb{h}^{t+1,n}\circ \theta').
\end{equation}
For LSTM, the situation becomes complicated due to the interaction between gates. Specifically, we can similarly have
\begin{equation}
\label{equ:LSTM_bp1}
\delta \pmb{h}^{t,n}=\frac{\partial \pmb{h}^{t,n+1}}{\partial \pmb{h}^{t,n}} \delta \pmb{h}^{t,n+1} + \frac{\partial \pmb{h}^{t+1,n}}{\partial \pmb{h}^{t,n}} \delta \pmb{h}^{t+1,n}
\end{equation}
where the two items on the right side represent the spatial gradient backpropagation and the temporal gradient backpropagation, respectively. Moreover, we can yield
\begin{equation}
\label{equ:LSTM_bp2}
\begin{cases}
\frac{\partial \pmb{h}^{t,n+1}}{\partial \pmb{h}^{t,n}}&=(\pmb{W}_{o1}^{n+1})^Tdiag(\theta_c(\pmb{c}^{t,n+1})\circ \sigma_o')\\
&+~(\pmb{W}_{f1}^{n+1})^Tdiag(\pmb{o}^{t,n+1}\circ \sigma_c' \circ \pmb{c}^{t-1,n+1} \circ \sigma_f') \\ &+~(\pmb{W}_{i1}^{n+1})^Tdiag(\pmb{o}^{t,n+1}\circ \sigma_c' \circ \pmb{g}^{t,n+1} \circ \sigma_i')\\
&+~(\pmb{W}_{g1}^{n+1})^Tdiag(\pmb{o}^{t,n+1}\circ \sigma_c' \circ \pmb{i}^{t,n+1} \circ \theta_g')\\
\frac{\partial \pmb{h}^{t+1,n}}{\partial \pmb{h}^{t,n}}&=(\pmb{W}_{o2}^{n})^Tdiag(\theta_c(\pmb{c}^{t+1,n})\circ \sigma_o')\\
&+~(\pmb{W}_{f2}^{n})^Tdiag(\pmb{o}^{t+1,n}\circ \sigma_c' \circ \pmb{c}^{t,n} \circ \sigma_f') \\ &+~(\pmb{W}_{i2}^{n})^Tdiag(\pmb{o}^{t+1,n}\circ \sigma_c' \circ \pmb{g}^{t+1,n} \circ \sigma_i')\\
&+~(\pmb{W}_{g2}^{n})^Tdiag(\pmb{o}^{t+1,n}\circ \sigma_c' \circ \pmb{i}^{t+1,n} \circ \theta_g')\\
\end{cases}
\end{equation}
where $diag(\cdot)$ converts a vector into a diagonal matrix.

\subsection{Rethinking the Differences}

Although SNNs, vanilla RNNs, and LSTM are similar in terms of information propagation paths, they are still quite different. In this subsection, we give our rethinking on their differences.

\begin{figure}[!htbp]
\centering     
\includegraphics[width=0.43\textwidth]{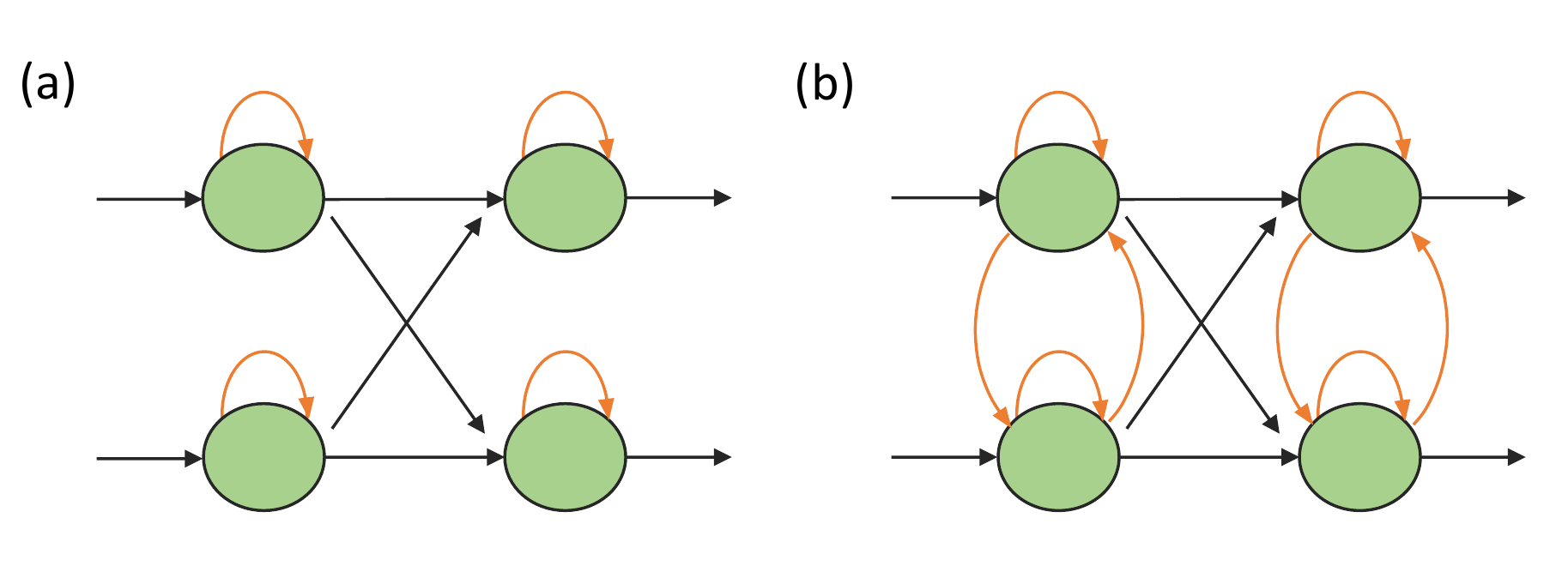}
\flushleft
\caption{\textbf{Connection pattern of (a) SNNs and (b) RNNs.}} \label{fig:connection} 
\end{figure}

\begin{figure}[!htbp]
\centering     
\includegraphics[width=0.45\textwidth]{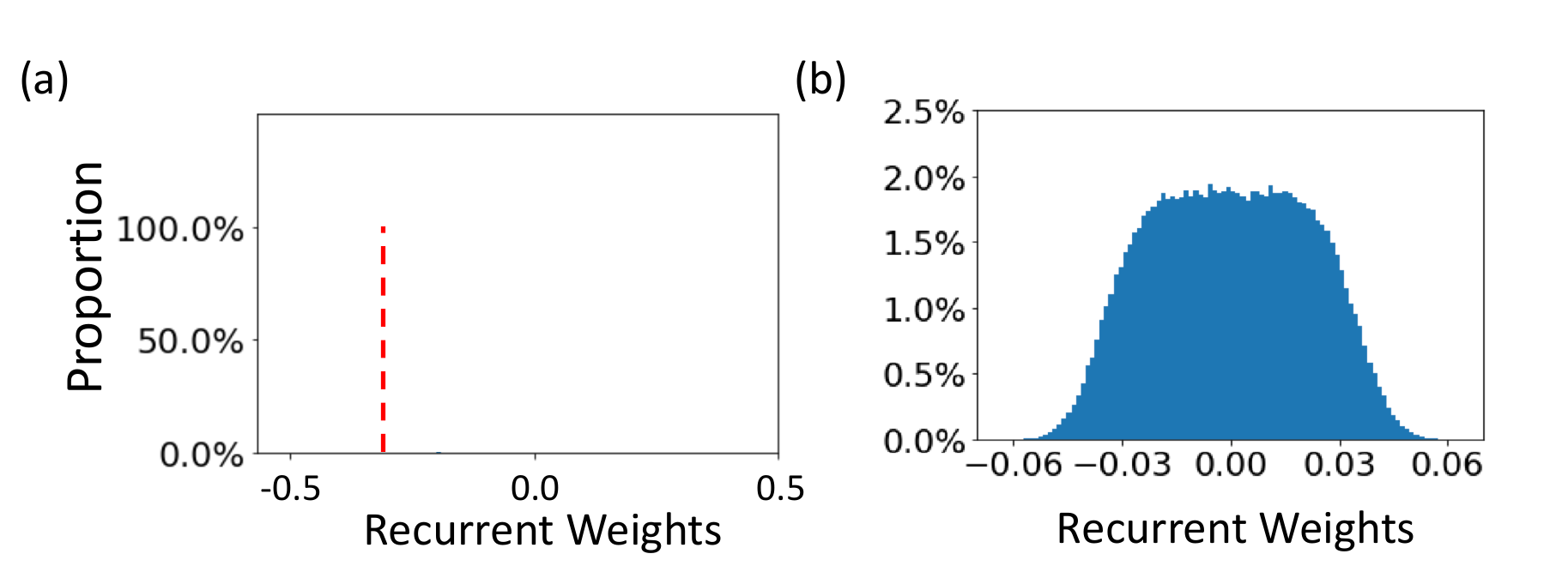}
\flushleft
\caption{\textbf{Distribution of recurrent weights in (a) SNNs and (b) RNNs.}} \label{fig:recurrent_weight}
\end{figure}

\textbf{Connection Pattern.} From Equation (\ref{equ:vanilla_RNN_layer})-(\ref{equ:SNN_layer_2}), it can be seen that the connection pattern of these models are different. First, for the neurons in the same layer, SNNs only have self-recurrence within each neuron, while RNNs have cross-recurrence among neurons. Specifically, the self-recurrence means that there are only intra-neuron recurrent connections; by contrast, the cross-recurrence allows inter-neuron recurrent connections within each layer. Second, the recurrent weights of SNNs are determined by the leakage factor of the membrane potential, which is restricted at $-e^{-\frac{dt}{\tau}}$; while in RNNs, the recurrent weights are trainable parameters. To make them clear, we use Figure \ref{fig:connection} to visualize the connection pattern of SNNs and RNNs and use Figure \ref{fig:recurrent_weight} to show the distribution of recurrent weights collected from practical models, which reflect the theoretical analysis.

\begin{figure}[!htbp]
\centering     
\includegraphics[width=0.485\textwidth]{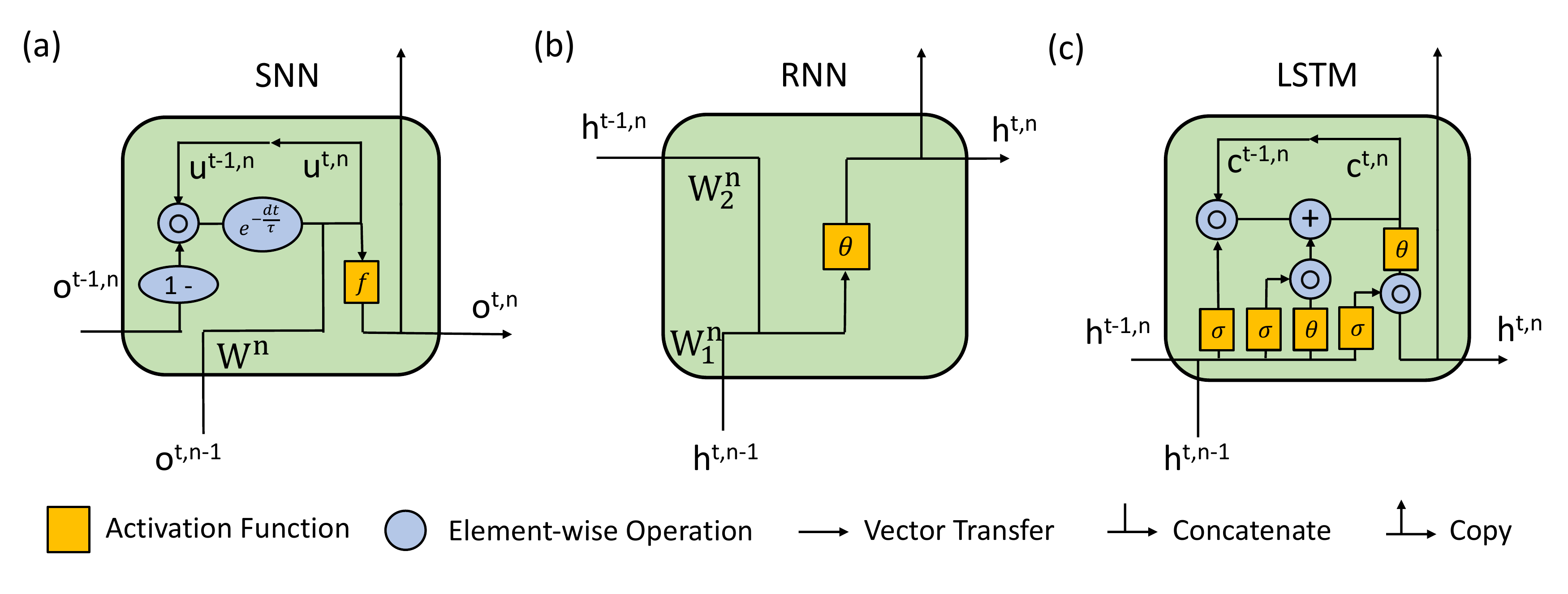}
\flushleft
\caption{\textbf{Neuron unit of (a) SNNs, (b) vanilla RNNs, and (c) LSTM.}} \label{fig:neuron} 
\end{figure}

\begin{figure}[!htbp]
\centering     
\includegraphics[width=0.485\textwidth]{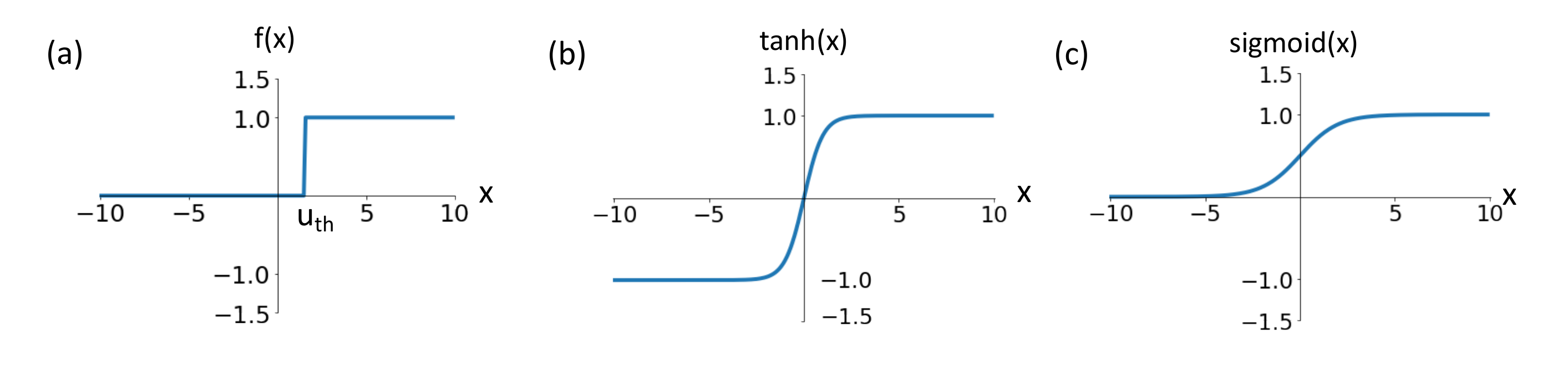}
\flushleft
\caption{\textbf{Activation function of (a) firing in SNNs, and (b) $\pmb{tanh}$ and (c) $\pmb{sigmoid}$ in RNNs.}} \label{fig:act_fun} 
\end{figure}

\textbf{Neuron Model.} Besides the analysis of the connection pattern, we discuss the modeling details inside each neuron unit. As depicted in Figure \ref{fig:neuron}, apparently, there are no gates in vanilla RNNs, unlike the complex gates in LSTM. For SNNs, as aforementioned, the extra membrane potential path is similar to the forget gate of LSTM; however, the reset mechanism bounds the membrane potential, unlike the unbounded cellular state in LSTM. In addition, as Figure \ref{fig:act_fun} shows, the activation function of SNNs is a firing function, which is essentially a step function with binary outputs; while the activation functions in vanilla RNNs and LSTM are continuous functions such as $tanh$ and $sigmoid$.  

\begin{table*}[!htbp]
\caption{Summary of similarities and differences among SNNs, vanilla RNNs, and LSTM.}
\label{tab:simlarties_differences}
\vspace{2pt}
\centering
\renewcommand\arraystretch{1.5}
\resizebox{0.985\textwidth}{!}{
\begin{tabular}{c | c | c | c | c | c | c }
\hline \hline
\multirow{2}*{\textbf{Model}} & \textbf{Spatiotemporal} & \textbf{Recurrence} & \textbf{Recurrent} & \multirow{2}*{\textbf{Gate Structure}} & \multirow{2}*{\textbf{Activation Function}} & \multirow{2}*{\textbf{Loss Function}} \\
 & \textbf{Path} & \textbf{Pattern} & \textbf{Weights} & & &  \\
\hline
SNNs & \Checkmark & Self-Neuron & $-e^{-\frac{dt}{\tau}}$ & Forget Gate & Binary: $fire$ ($f$) & $L = ||\pmb{Y}^{label}-\frac{1}{T}\sum_{t=1}^{t}\pmb{o}^{t,N}||_{2}^{2}$ \\
Vanilla RNNs & \Checkmark & Cross-Neuron & Trainable & \XSolidBrush & Continuous: $tanh$ ($\theta$) & Flexible \\
LSTM & \Checkmark & Cross-Neuron & Trainable & Multiple Gates & Continuous: $sigmoid$ ($\sigma$) \& $tanh$ ($\theta$) & Flexible \\
\hline \hline
\end{tabular}}
\end{table*}

\textbf{Loss Function.} Under the framework of gradient descent based supervised learning, a loss function is critical for the overall optimization. The loss function formats for SNNs and RNNs are different. Specifically, for SNNs, the spike rate coding scheme is usually combined with the mean square error (MSE) to form a loss function, which can be abstracted as
\begin{equation}
\label{equ:SNN_loss}
L = ||\pmb{Y}^{label}-\frac{1}{T}\sum_{t=1}^{t}\pmb{o}^{t,N}||_{2}^{2}
\end{equation}
where $\pmb{Y}^{label}$ is the label, $\pmb{o}^{t,N}$ is the output of the last layer, and $T$ is the number of simulation timesteps during training. This loss function takes the output spikes fired at all timesteps into account, and thus the neuron fires the most determines the recognition result. Different from Equation (\ref{equ:SNN_loss}), the mainstream loss function for RNNs usually obeys
\begin{equation}
\label{equ:RNN_loss1}
L = ||\pmb{Y}^{label}-\pmb{W}^y\pmb{h}^{T,N}||_{2}^{2}
\end{equation}
where $\pmb{h}^{T,N}$ is the hidden state of the last layer at the last timestep and $\pmb{W}^y$ is a trainable weight matrix.

\subsection{Comparison Methodology}

Based on the above analysis, we summarize the similarities and differences among SNNs, vanilla RNNs, and LSTM in Table \ref{tab:simlarties_differences}. Owing to the similar spatiotemporal dynamics, it is possible to benchmark all these models on neuromorphic datasets. Moreover, facing the differences, we appropriately unify the following aspects to ensure comparability and fairness in our evaluation. 

\bigskip
\subsubsection{Dataset Selection and Temporal Resolution Tuning}\quad

We benchmark all models on two neuromorphic vision datasets: one is a DVS-converted dataset named N-MNIST and the other is a DVS-captured dataset named DVS Gesture, which are already introduced in Section \ref{sec:preliminary:dataset}. The detailed information of the two selected datasets is provided in Table \ref{tab:dataset}. For SNNs, the processing of neuromorphic data is natural due to the same spatiotemporal components and event-driven fashion; while for RNNs, the spike data are just treated as binary values, i.e. $\{0,~1\}$. 

\begin{table}[!htbp]
\caption{Details of the selected neuromorphic datasets.}
\label{tab:dataset}
\vspace{2pt}
\centering
\begin{tabular}{ccc}
\hline\hline
\textbf{Dataset}     & N-MNIST  & DVS Gesture      \\ \hline
\textbf{Description} & Handwritten Digits & Human Gestures \\
\textbf{Slice Size}  & 34$\times$34  & 128$\times$128  \\
$dt_0$       & $1\mu s$               & $1\mu s$ \\
\textbf{\#Training Slices} & 60000    & 1176   \\
\textbf{\#Testing Slices}  & 10000    & 288     \\ 
\hline\hline
\end{tabular}
\end{table}

\begin{figure}[!htbp]
\centering     
\includegraphics[width=0.48\textwidth]{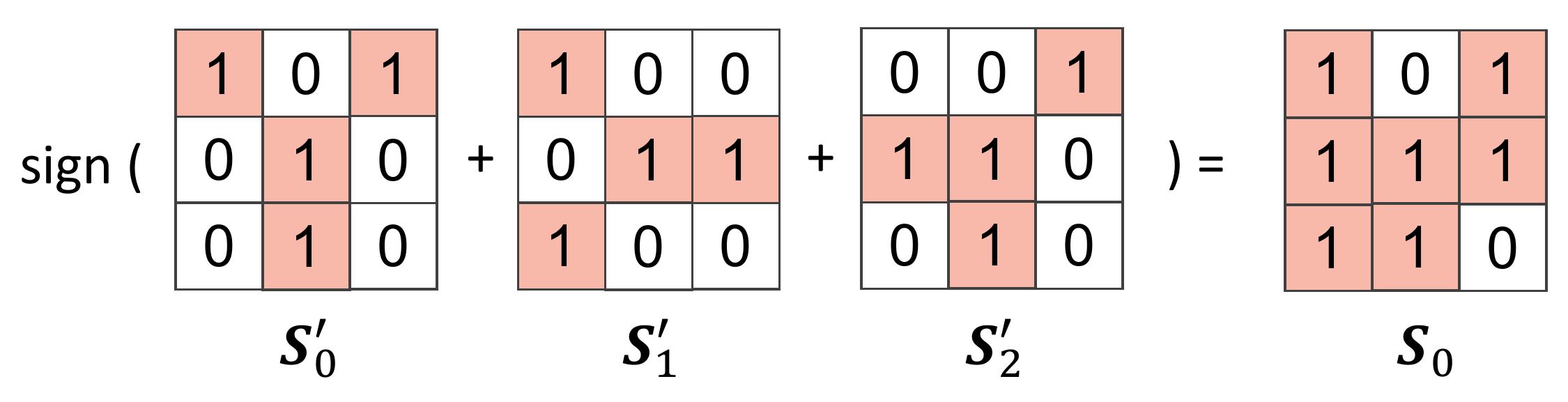}
\flushleft
\caption{\textbf{Illustration of temporal collapse for tunable temporal resolution.}} \label{fig:collapse} 
\end{figure}

Usually, the original recording time length of each spike pattern is very long, e.g. $10^5$. The underlying reason is due to the fine-grained temporal resolution of DVS cameras, originally at $\mu s$ level. However, the simulation timestep number of neural networks cannot be too large, otherwise, the time and memory costs during training are unaffordable. To this end, we consider the flexibility in tuning the temporal resolution. Specifically, every multiple successive slices of spike events in the original dataset within each temporal resolution unit are collapsed along the temporal dimension into one slice. Here the temporal collapse means there will be a spike at the resulting pixel if there exist spikes at the same location in any original slices within the collapse window. We describe the collapse process as
\begin{equation}
\label{equ:collapse}
\pmb{S}_t= sign(\sum_{t'} \pmb{S}'_{t'}),~s.t.~~t'\in [\alpha_{dt}\times t,~\alpha_{dt}\times(t+1)-1]
\end{equation}
where $\pmb{S}'$ denotes the original slice sequence, $t'$ is the original recording timestep index, $\pmb{S}$ denotes the new slice sequence after collapse, $t$ is the new recording timestep index, and $\alpha_{dt}$ is the temporal resolution factor. $sign$ is defined as: $sign(x)=1,~if~x>0$; $sign(x)=0,~if~x=0$; $sign(x)<0,~if~x<0$. After collapse, the original slice sequence $\{\pmb{S}'_{t'},~t'\in [0,~T_0-1]\}$ becomes a new slice sequence  $\{\pmb{S}_{t},~t\in [0,~T_0/\alpha_{dt}-1]\}$. Apparently, the actual temporal resolution ($dt$) satisfies
\begin{equation}
\label{equ:tr}
dt=\alpha_{dt}dt_0
\end{equation}
where $dt_0$ is the original temporal resolution. Figure \ref{fig:collapse} illustrates an example of temporal collapse with $\alpha_{dt}=3$. 

\begin{figure}[!htbp]
\centering     
\includegraphics[width=0.32\textwidth]{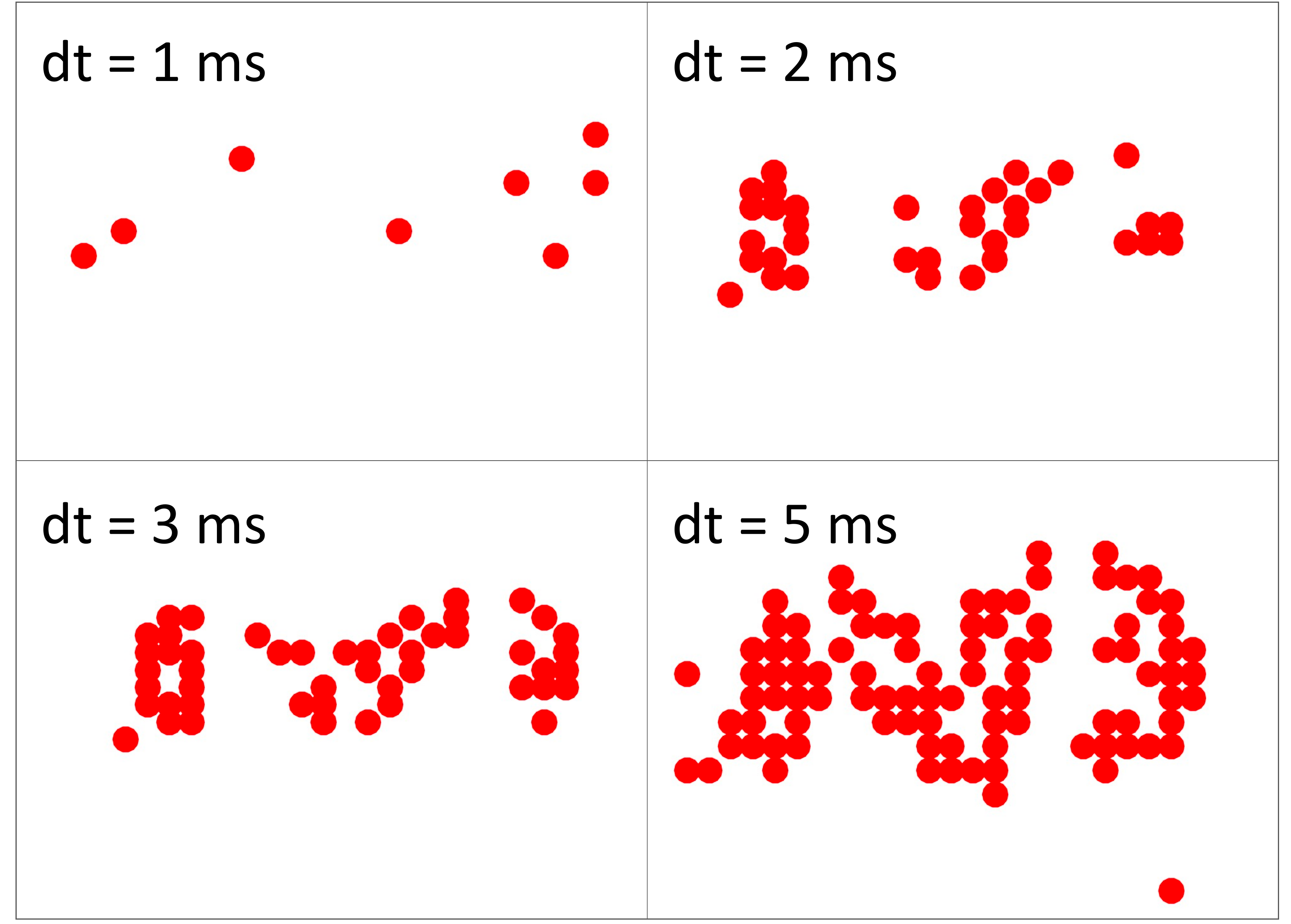}
\caption{\textbf{Slices of N-MNIST dataset under different temporal resolution, taking the digit ``3'' for example.}}
\label{fig:NMNIST_dt} 
\end{figure}

A large temporal resolution will increase the spike rate of new slices, as demonstrated in Figure \ref{fig:NMNIST_dt}. In addition, at each simulation timestep in Equation (\ref{equ:vanilla_RNN_layer})-(\ref{equ:SNN_layer_2}), the neural network processes one slice after the temporal collapse. Therefore, if the simulation timestep number $T$ remains fixed, a larger temporal resolution could extend the actual simulation time, which is able to capture more temporal dependence in the neuromorphic dataset. By tuning the temporal resolution, we create opportunities to extract more insights from the change of model performance.

\bigskip
\subsubsection{Learning Algorithm and Loss Function Design}\quad

Since RNNs are normally trained by supervised BPTT algorithm, to make the comparison fair, we select a recent BPTT-inspired learning algorithm with spatiotemporal gradient propagation for SNNs \cite{wu2018spatio}. Regarding the loss function required by gradient descent, the one in Equation (\ref{equ:SNN_loss}) based on the rate coding scheme and MSE is widely used for SNNs. Under this loss function, there is gradient feedback at every timestep, which can alleviate the gradient vanishing problem to some extent.

In contrast, the existing loss functions for RNNs are flexible, including the mainstream one shown in Equation (\ref{equ:RNN_loss1}) that considers only the output at the last timestep and others that consider the outputs at all timesteps \cite{werbos1990backpropagation,boden2002guide,vlachas2020backpropagation} such as 
\begin{equation}
\label{equ:RNN_loss2}
L = \frac{1}{T}\sum_t ||\pmb{Y}_t^{label} - \pmb{W}^y\pmb{h}^{t,N}||_{2}^{2}.
\end{equation}
However, even if using the above loss function that considers the outputs at all timesteps, it is still slightly different from the one in Equation (\ref{equ:SNN_loss}) for SNNs. To make the comparison fair, we provide two kinds of loss function configuration for RNNs. One is the mainstream loss function as in Equation (\ref{equ:RNN_loss1}); the other is a modified version of Equation (\ref{equ:RNN_loss2}), i.e.,
\begin{equation}
\label{equ:RNN_loss3}
L = ||\pmb{Y}^{label}-\frac{1}{T}\sum_t (\pmb{W}^y\pmb{h}^{t,N})||_{2}^{2}.
\end{equation}
For clarity, we term the above format in Equation (\ref{equ:RNN_loss3}) for RNNs as the rate-coding-inspired loss function.

\bigskip
\subsubsection{Network Structure and Hyper-parameter Setting}\quad

The FC layer based structure is widely used in SNNs and RNNs, which is termed as multilayered perceptron (MLP) based structure in this work. Whereas, the learning performance of MLP-based structures is usually poor, especially for visual tasks. To this end, the Conv layer based structure is introduced into SNNs to improve the learning performance \cite{wu2019direct}, which is termed as convolutional neural network (CNN) based structure in this work. Facing this situation, besides the basic MLP structure, we also implement the CNN structure for RNNs, including both vanilla RNNs and LSTM. In this way, the comparison between different models is restricted on the same network structure, which is more fair. Table \ref{tab:net_structure} provides the network structure configuration on different datasets. Since N-MNIST is a simpler task, we only use the MLP structure; while for DVS Gesture, we adopt both MLP and CNN structures.

\begin{table}[!htbp]
\caption{Network structure configuration.}
\label{tab:net_structure}
\vspace{2pt}
\centering
\renewcommand\arraystretch{1.3}
\resizebox{0.45\textwidth}{!}{
\begin{tabular}{c | c }
\hline\hline
\textbf{Neuromorphic Vision Dataset}   & \textbf{Network Structure} \\ \hline
N-MNIST                     & MLP: Input-512FC-10              \\ \hline
\multirow{3}*{DVS Gesture} & MLP: Input-MP4-512FC-11               \\
                            & CNN: Input-MP4-64C3-128C3-\\
                            &AP2-128C3-AP2-256FC-11 \\ 
                            \hline\hline
\end{tabular}}\\
 \begin{tablenotes}
\item[1] \footnotesize{\small Note: $n$C3-Conv layer with $n$ output feature maps and $3\times 3$ weight kernel size, MP4-max pooling with $4\times 4$ pooling kernel size, AP2-average pooling with $2\times 2$ pooling kernel size.}
\end{tablenotes}
\end{table}

Besides the network structure, the training process needs some hyper-parameters such as number of epochs, number of timesteps, batch size, learning rate, etc. To ensure fairness, we unify the training hyper-parameters of different models. Specifically, as listed in Table \ref{tab:para_setting}, except for the unique hyper-parameters for SNNs, others are shared by all models.

\begin{table}[!htbp]
\caption{Hyper-parameter setting.}
\label{tab:para_setting}
\vspace{2pt}
\centering
\renewcommand\arraystretch{1.3}
\resizebox{0.485\textwidth}{!}{
\begin{tabular}{c c c c c c}
\hline\hline
\textbf{Model} & \textbf{Hyper-parameter}  & \textbf{Description}   & \textbf{N-MNIST} & \multicolumn{2}{c}{\textbf{DVS Gesture}} \\ \hline
\multirow{4}*{Shared} & Max Epoch   & -   & 100     & 100            & 100           \\
& Batch Size           & -                      & 50      & 36             & 36            \\
& $T $              & Timestep Number & 15      & 60             & 60            \\
& lr                   & Learning Rate          & $1e^{-4}$    & $1e^{-4}$            & $1e^{-4}$           \\ 
\hline
\multirow{3}*{SNNs} & $u_{th}$          & Firing Threshold       & 0.3     & 0.3            & 0.3           \\
& $e^{-\frac{dt}{\tau}}$  & Leakage Factor       & 0.3     & 0.3             & 0.3          \\
& a                    & Gradient Width         & 0.25    & 0.25           & 0.5           \\\hline\hline
\end{tabular}}
\end{table}

In summary, with the above rethinking on the similarities and differences as well as the proposed solutions, we successfully unify several aspects involving testing datasets, temporal resolution, learning algorithm, loss function, network structure, hyper-parameter, etc., which are listed in Table \ref{tab:unification}. This unification ensures the comparability between SNNs and RNNs, and further makes the comparison fair, which lays the foundation of this work. 

\begin{table}[!htbp]
\caption{Unification for comparison.}
\label{tab:unification}
\vspace{2pt}
\centering
\renewcommand\arraystretch{1.4}
\resizebox{0.49\textwidth}{!}{
\begin{tabular}{c | c }
\hline\hline
\textbf{Neuromorphic Vision Dataset}        & N-MNIST \& DVS Gesture    \\ 
\textbf{Temporal Resolution}         & Tunable ($dt$)  \\ 
\textbf{Learning Algorithm}         & BPTT-inspired (SNNs); BPTT (RNNs)  \\ 
\textbf{Loss Function}         & Rate Coding (SNNs); Mainstream or Rate-Coding-Inspired (RNNs)  \\ 
\textbf{Network Structure}     & MLP \& CNN \\
\textbf{Hyper-parameter}       & SNN-Specialized \& Shared  \\
\hline\hline
\end{tabular}}
\end{table}

%% file: text/Results.tex
\section{Results}\label{sec:result}

With the unification mentioned in Table \ref{tab:unification}, we conduct a series of contrast experiments and extract some insights in this section.

\subsection{Experimental Setup}
All models are implemented in the open-source framework, Pytorch \cite{paszke2019pytorch}. The configurations of network structures and training hyper-parameters are already provided in Table \ref{tab:net_structure} and Table \ref{tab:para_setting}, respectively. The temporal resolution has six levels on N-MNIST ($dt=\{1ms,~2ms,~3ms,~5ms,~10ms,~20ms\}$) and other six levels on DVS Gesture ($dt=\{1ms,~5ms,~10ms,~15ms,~20ms,~25ms\}$). With a given number of simulation timesteps (i.e. $T$), it means only the first $T$ slices of each spike pattern will be used during training and testing. Usually, we fix the $T$ value; while in Section \ref{sec:generalization}, we fix the simulation temporal length (i.e. $T\times dt$) rather than $T$. Unless otherwise specified, the leakage factor (i.e. $e^{-\frac{dt}{\tau}}$) is fixed at 0.3. In addition, the Adam (adaptive moment estimation) optimizer \cite{kingma2014adam} with the default parameter setting ($\alpha=1e^{-4}$, $\beta_{1}=0.9$, $\beta_{2}=0.999$, $\epsilon=1e^{-8}$) is used for the adjustment of network parameters. 

\begin{table*}[!htbp]
\caption{Accuracy of MLP-based models on N-MNIST.}
\label{tab:acc_NMNIST}
\vspace{2pt}
\centering
\renewcommand\arraystretch{1.3}
\resizebox{0.7\textwidth}{!}{
\begin{tabular}{cccccc}
\hline\hline
 \multirow{2}*{$dt$}   & \multirow{2}*{\textbf{SNNs}}   & \multirow{2}*{\textbf{Vanilla RNNs}}  & \multirow{2}*{\textbf{LSTM}} & \textbf{Vanilla RNNs} & \textbf{LSTM} \\ 
 & & & & \textbf{(Rate-Coding-inspired)} & \textbf{(Rate-Coding-inspired)} \\ \hline
$1ms$ & 84.96\% & 66.74\% & 65.33\% & 87.62\% & 82.25\% \\
$2ms$ & 95.94\% & 90.54\% & 93.18\% & 95.97\% & 97.08\%  \\
$3ms$ & 98.19\% & 95.41\% & 96.70\% & 98.15\% & 98.69\%   \\
$5ms$ & 98.28\% & 92.37\% & 94.24\% & 98.58\% & 98.68\%  \\ 
\hline\hline
\end{tabular}}
\end{table*}

\begin{table*}[!htbp]
\caption{Accuracy of MLP-based models on DVS Gesture.}
\label{tab:acc_MLP_DVSGuesture}
\vspace{2pt}
\centering
\renewcommand\arraystretch{1.3}
\resizebox{0.7\textwidth}{!}{
\begin{tabular}{cccccc}
\hline\hline
 \multirow{2}*{$dt$}   & \multirow{2}*{\textbf{SNNs}}   & \multirow{2}*{\textbf{Vanilla RNNs}}  & \multirow{2}*{\textbf{LSTM}} & \textbf{Vanilla RNNs} & \textbf{LSTM} \\ 
 & & & & \textbf{(Rate-Coding-inspired)} & \textbf{(Rate-Coding-inspired)} \\ \hline
$1ms$ & 54.51\% &16.32\% & 19.79\% & 19.44\%   &  50.35\% \\
$5ms$ & 76.04\% & 27.78\% & 45.49\% & 30.90\%  & 74.31\%  \\
$10ms$ & 82.63\% & 30.56\% & 37.15\% & 36.11\%  & 84.03\%  \\
$15ms$ & 85.07\% & 33.68\% & 46.88\% & 42.01\%  & 85.42\%  \\
$20ms$ & 86.81\% & 38.54\% & 42.71\% & 52.78\%  & 88.19\%  \\
$25ms$ & 87.50\% & 19.44\% & 44.79\% & 50.35\%  & 86.81\%  \\ 
\hline\hline
\end{tabular}}
\end{table*}

\begin{table*}[!htbp]
\caption{Accuracy of CNN-based models on DVS Gesture.}
\label{tab:acc_CNN_DVSGuesture}
\vspace{2pt}
\centering
\renewcommand\arraystretch{1.3}
\resizebox{0.7\textwidth}{!}{
\begin{tabular}{cccccc}
\hline\hline
 \multirow{2}*{$dt$}   & \multirow{2}*{\textbf{SNNs}}   & \multirow{2}*{\textbf{Vanilla RNNs}}  & \multirow{2}*{\textbf{LSTM}} & \textbf{Vanilla RNNs} & \textbf{LSTM} \\ 
 & & & & \textbf{(Rate-Coding-inspired)} & \textbf{(Rate-Coding-inspired)} \\ \hline
$1ms$ & 71.53\% & 48.26\% & 64.58\% & 56.25\%  & 65.63\%  \\
$5ms$ & 87.15\% & 51.74\% & 79.86\% & 75.35\%   & 85.76\%  \\
$10ms$ & 91.67\% & 56.94\% & 84.72\% & 82.99\%   & 86.81\%   \\
$15ms$ & 93.05\% & 58.68\% & 91.67\% & 84.02\%   & 93.40\%  \\
$20ms$ & 92.71\% & 65.97\% & 89.24\% & 90.27\%   & 92.70\%  \\
$25ms$ & 93.40\% & 70.49\% & 92.36\% & 92.01\%   & 93.75\%  \\ 
\hline\hline
\end{tabular}}
\end{table*}

\begin{figure*}[!htbp]
\centering     
\includegraphics[width=0.85\textwidth]{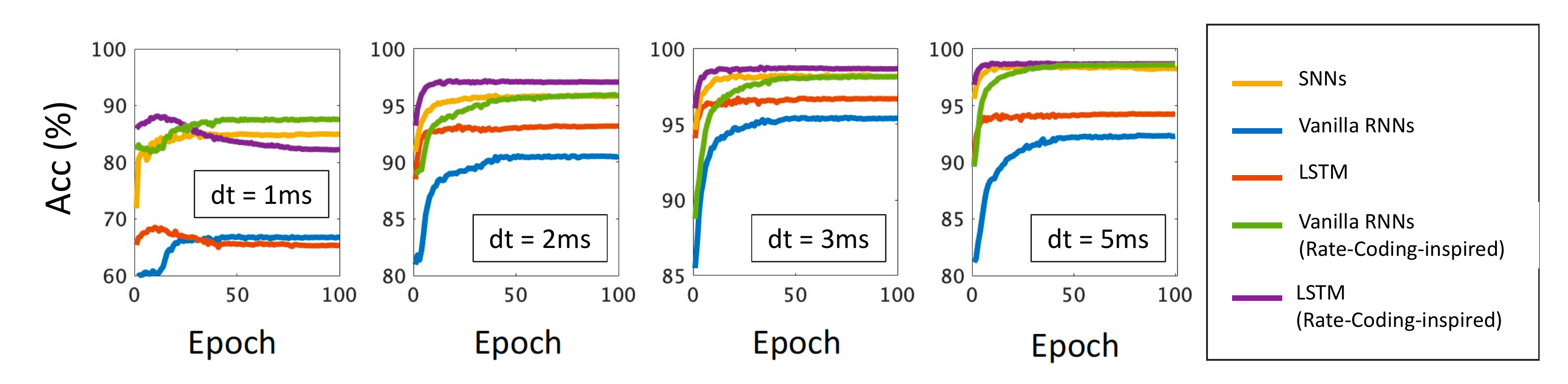}
\caption{\textbf{Training curves of different models under different $dt$ settings on N-MNIST.}} \label{fig:training_curve} 
\end{figure*}

\subsection{Overall Accuracy Comparison}\label{sec:result_acc}

Tables \ref{tab:acc_NMNIST}-\ref{tab:acc_CNN_DVSGuesture} list the accuracy results of a series of contrast experiments on both N-MNIST and DVS Gesture datasets. On N-MNIST, SNNs achieve the best accuracy among the common models. Interestingly, when we apply the rate-coding-inspired loss function (Equation (\ref{equ:RNN_loss3})), RNNs can achieve comparable or even better accuracy than SNNs. A similar trend is also found in DVS Gesture. However, it seems that the vanilla RNNs cannot outperform SNNs on DVS Gesture, especially in the MLP-based cases, even if the rate-coding-inspired loss function is used. The underlying reason might be due to the gradient problem. As well known, compared to vanilla RNNS, LSTM can alleviate the gradient vanishing issue via the complex gate structure, thus achieving much longer temporal dependence \cite{hochreiter1997long,gers1999learning}. For SNNs, the membrane potential can directly impact the neuronal state at the next timestep, leading to one more information propagation path over vanilla RNNs in both forward and backward passes (see Figure \ref{fig:dataflow}). This extra path acts similarly as the LSTM's forget gate (i.e. the most important gate of LSTM \cite{jozefowicz2015empirical}), thus it can also memorize longer-term dependence than vanilla RNNs and improve accuracy.

Figure \ref{fig:training_curve} further presents the training curves of these models on N-MNIST. It could be observed that the common RNNs converge poorly on neuromorphic datasets while the RNNs with the rate-coding-inspired loss function can shift the training curves upward, which demonstrates of the effectiveness of the rate-coding-inspired loss function. Moreover, we find that SNNs and LSTM converge faster than vanilla RNNs. All these observations are consistent with the results in Tables \ref{tab:acc_NMNIST}-\ref{tab:acc_CNN_DVSGuesture}. 

\begin{figure}[!htbp]
\centering     
\includegraphics[width=0.49\textwidth]{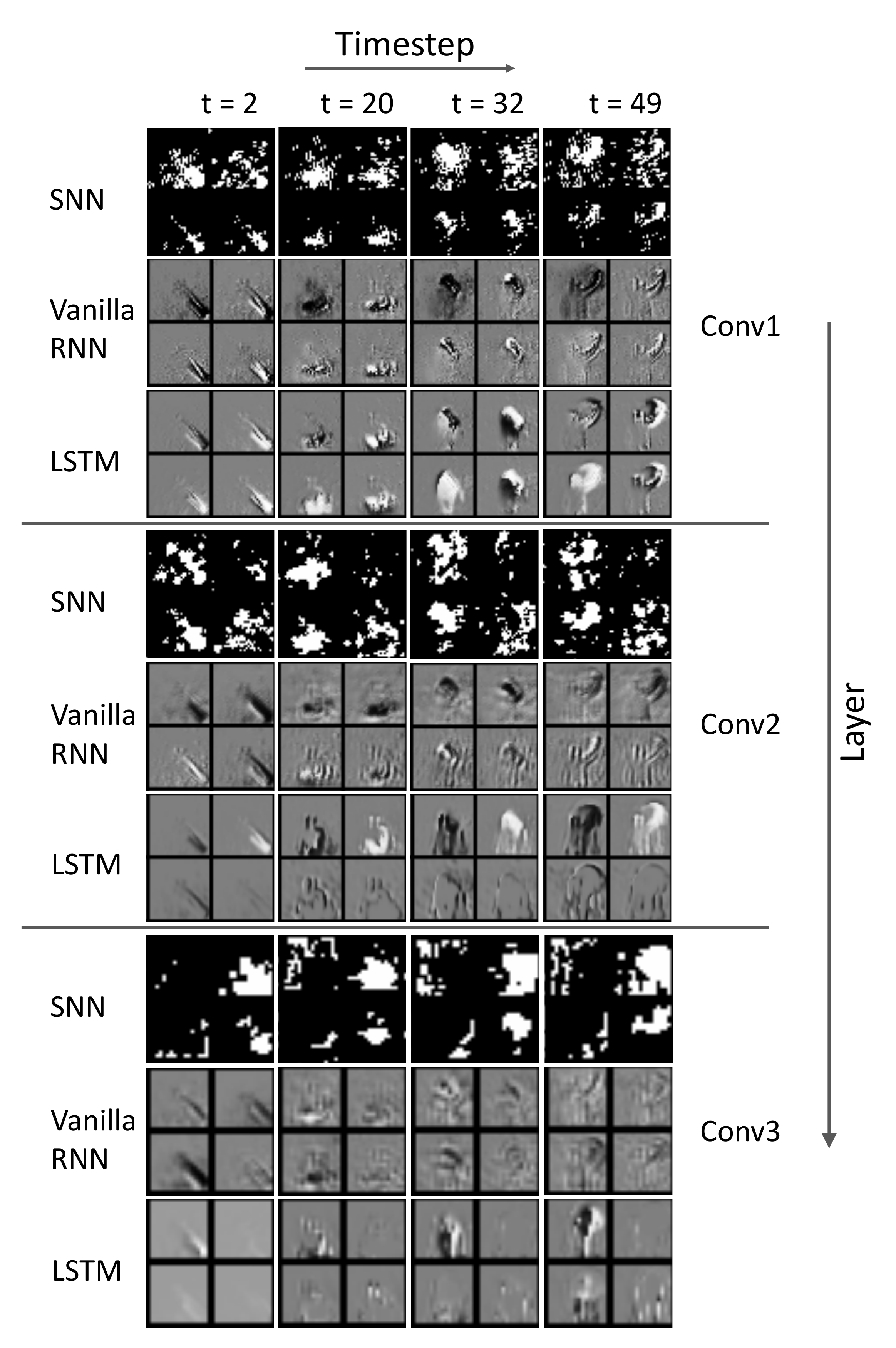}
\caption{\textbf{Visualization of feature maps across timesteps in different CNN-based models on DVS Gesture.} We set $dt=15ms$ and take the left arm clockwise gesture for example.} \label{fig:FM} 
\end{figure}

Besides the above analysis, we further visualize the feature maps of Conv layers on DVS Gesture to see what happens, as shown in Figure \ref{fig:FM}. For simplicity, here we only visualize the case of $dt=15ms$; the RNN models are improved by the rate-coding-inspired loss function. Among the three models, the vanilla RNN has the most clear feature maps, close to the input slices at the corresponding timesteps. However, the feature maps of SNN and LSTM models obviously include an integration of the current timestep and traces of previous timesteps, which owes to the extra membrane potential path of SNNs and the complex gate structure of LSTM. In the feature maps of SNNs and LSTM, the entire area passed by the dynamic gesture is lighted up, making them look like comets. This feature integration strengthens the temporal dependence, which further changes the later layers from learning temporal features to learning spatial features to some extent. On DVS-captured datasets like DVS Gesture, the different input slices across timesteps jointly constitute the final pattern to recognize; while on DVS-converted datasets like N-MNIST, the slices at different timesteps are close. This, along with the longer-term memory of SNNs and LSTM, can explain that the accuracy gap between vanilla RNNs and SNNs/LSTM is larger on DVS Gesture than that on N-MNIST.

\begin{table}[!htbp]
\caption{Influence of membrane potential leakage and reset. }
\label{tab:leak_reset}
\vspace{2pt}
\centering
\renewcommand\arraystretch{1.3}
\resizebox{0.25\textwidth}{!}{
\begin{tabular}{ccc}
\hline \hline
\textbf{Leakage}   & \textbf{Reset}   & \textbf{Accuracy}  \\ \hline
\Checkmark & \Checkmark & 98.05\% \\
\hline
\Checkmark & \XSolidBrush & 98.13\% \\
\XSolidBrush & \Checkmark & 97.66\% \\
\XSolidBrush & \XSolidBrush & 96.27\% \\
\hline\hline
\end{tabular}}
\end{table}

Furthermore, we do an extra experiment to investigate the influence of the membrane potential leakage and reset mechanisms for SNNs. Here we test on N-MNIST with $dt=7.5ms$. As presented in Table \ref{tab:leak_reset}, the removal of these mechanisms will degrade the accuracy. In fact, both leakage and reset can reduce the membrane potential, thus lowering the spike rate to some extent, which is helpful to improve the neuronal selectivity. Interestingly, we find the joint impact of the two mechanisms is larger than the impact of any one of them.

\begin{table}[!htbp]
\caption{Accuracy results of prior work on SNNs.}
\label{tab:prior_SNN}
\vspace{2pt}
\centering
\renewcommand\arraystretch{1.3}
\resizebox{0.48\textwidth}{!}{
\begin{tabular}{cccc}
\hline\hline
\textbf{Dataset}           & \textbf{Work}   & \textbf{Network Structure}  & \textbf{Accuracy} \\ \hline
\multirow{8}{*}{N-MNIST}    & SKIM \cite{cohen2016skimming}       & Input-800FC-10             & 92.87\%  \\
                              & Lee et al. \cite{lee2016training} & Input-10000FC-10           & 98.66\%  \\
                             & DART \cite{ramesh2019dart}       & DART Feature Descriptor    & 97.95\%  \\
                             & SLAYER \cite{shrestha2018slayer}     & Input-500FC-500FC-10         & 98.89\%  \\
                             & SLAYER \cite{shrestha2018slayer}     & Input-12C5-AP2-64C5-AP2-10 & 99.20\%  \\
                             & STBP \cite{wu2018spatio}&Input-800FC-10 & 98.78\%\\
                             & Wu et al. \cite{wu2019direct}&SNN (CNN-based 8 layers) & 99.53\%\\
                             & Ours       & Input-512FC-10             &  98.28\%        \\ \hline
\multirow{3}{*}{DVS Gesture} & TureNorth \cite{amir2017low}  & SNN (CNN-based 16 layers)            & 91.77\%  \\
                             & SLAYER \cite{shrestha2018slayer}     & SNN (CNN-based 8 layers)             & 93.64\%  \\
                             & Ours       & SNN (CNN-based 8 layers)  & 93.40\%\\ \hline\hline
\end{tabular}}
\end{table}

At last, we provide the accuracy results of several prior works that applied SNNs on the two neuromorphic datasets. Note that we do not provide results involving RNNs since rare work tested them on neuromorphic datasets. As depicted in Table \ref{tab:prior_SNN}, our SNN models can achieve acceptable results, although not the best. Since our focus is the comparison between SNNs and RNNs rather than beating prior work, we do not adopt large models and complex optimization strategies used in prior work to improve accuracy.

\subsection{Temporal Resolution Analysis}\label{sec:generalization}

Also from Tables \ref{tab:acc_NMNIST}-\ref{tab:acc_CNN_DVSGuesture}, we find that as the temporal resolution grows larger, the accuracy will be improved. The reasons are two-fold: on the one hand, the spike events become dense when $dt$ lies in large values (see Figure \ref{fig:NMNIST_dt}), which usually forms more effectual features in each slice; on the other hand, with the same number of simulation timesteps (i.e. $T$) during training, a larger temporal resolution can include more slices in the original dataset, which provides more information of the moving object. Furthermore, we find that SNNs achieve significant accuracy superiority on DVS-captured datasets like DVS Gesture when the temporal resolution is small (e.g. $dt\leq 10ms$). This indicates that, unlike the continuous RNNs, the event-driven SNNs are more suitable to extract sparse features, which is also pointed out in \cite{deng2020rethinking}. On DVS-converted datasets like N-MNIST, the sparsity gap of spike events under different temporal resolution is usually smaller than that on DVS-captured datasets, thus the accuracy superiority of SNNs is degraded.

\begin{table}[!htbp]
\caption{Influence of temporal resolution under the same simulation temporal length.}
\label{tab:influence_dt}
\vspace{2pt}
\centering
\renewcommand\arraystretch{1.3}
\resizebox{0.485\textwidth}{!}{
\begin{tabular}{cccc}
\hline \hline
 \multirow{2}*{$dt$}   & \multirow{2}*{\textbf{SNNs}}  & \textbf{Vanilla RNNs} & \textbf{LSTM} \\ 
 & & \textbf{(Rate-Coding-inspired)} & \textbf{(Rate-Coding-inspired)} \\ \hline
$5ms$   & 97.48\% & 97.97\% & 98.48\% \\
$10ms$   & 96.99\% & 98.13\% & 98.45\% \\
$20ms$   & 96.59\% & 97.47\% & 97.83\% \\
\hline\hline
\end{tabular}}
\end{table}

In essence, the influence of temporal resolution increase is not always positive. As illustrated in Figure \ref{fig:collapse}, the temporal collapse as $dt$ grows also loses some spikes, leading to temporal precision loss. To investigate the negative effect, we conduct experiments on N-MNIST with large $dt$ values. To eliminate the impact of different simulation temporal length (i.e. $T\times dt$) when $dt$ varies, we adapt the number of simulation timesteps to ensure the same simulation temporal length, i.e. fixing $T\times dt=40ms$ here. The results are given in Table \ref{tab:influence_dt}. As $dt$ excessively increases, the accuracy degrades due to the temporal precision loss.

\begin{table}[!htbp]
\caption{Accuracy results of generalization test under the same simulation temporal length but different temporal resolution.}
\label{tab:acc_generalization}
\vspace{2pt}
\centering
\renewcommand\arraystretch{1.3}
\resizebox{0.485\textwidth}{!}{
\begin{tabular}{cccc}
\hline \hline
$dt$ \textbf{during Testing} & $3ms$   & $2ms$   & $1ms$   \\ \hline
\textbf{SNN}                      & 98.19\% & 97.31\% & 96.01\% \\
\textbf{SNN (Adaptive Leakage)}   & 98.19\% & 97.83\% & 97.04\% \\
\textbf{SNN (Cross-Recurrence)}   & 98.32\% & 97.62\% & 73.51\% \\
\textbf{Vanilla RNN (Rate-Coding-inspired)}   & 98.15\% & 97.09\% & 78.33\% \\
\textbf{LSTM (Rate-Coding-inspired)}   & 98.69\% & 97.98\% & 77.82\% \\
\hline\hline
\end{tabular}}
\end{table}

\begin{figure}[!htbp]
\centering     
\includegraphics[width=0.485\textwidth]{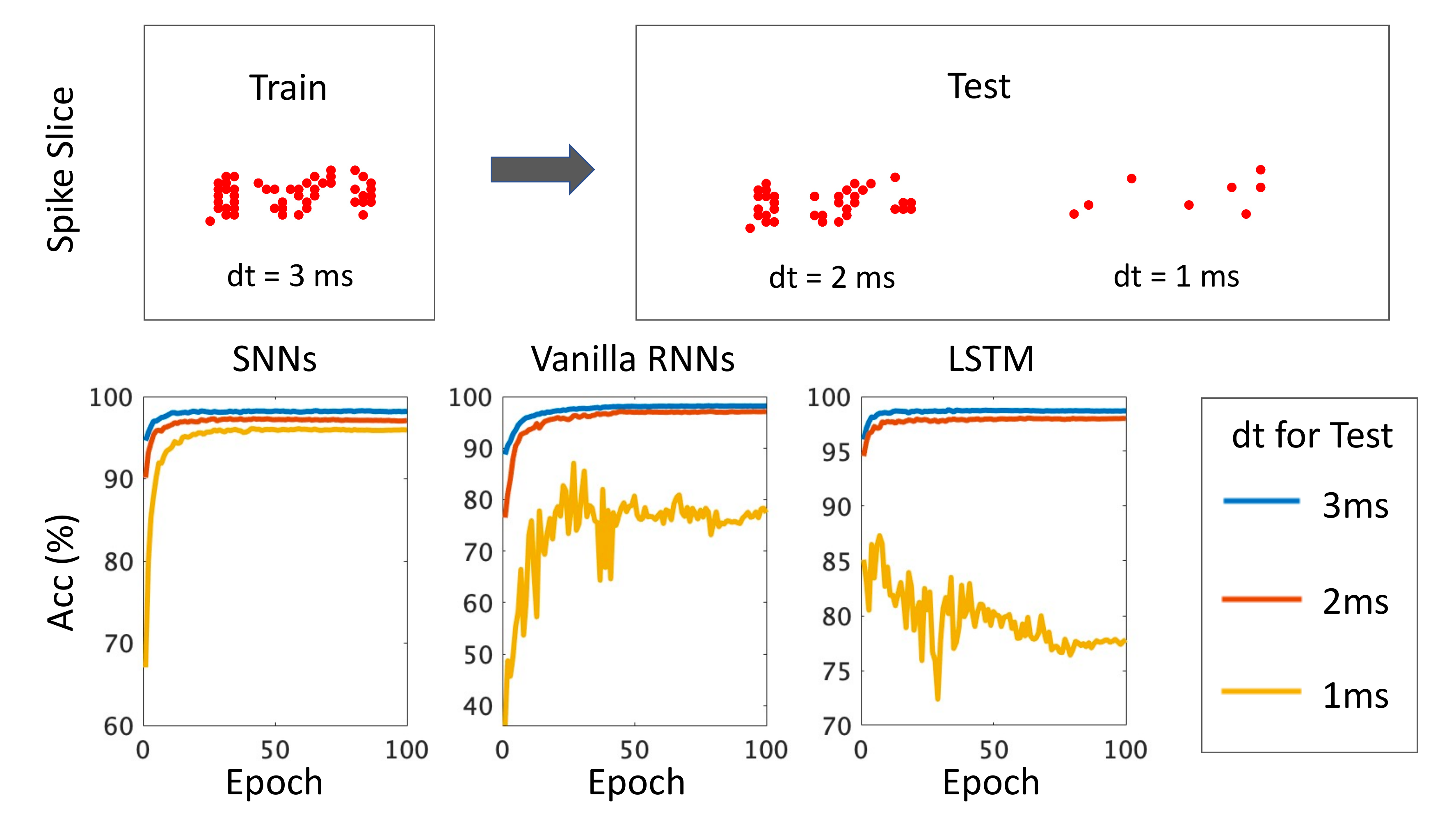}
\caption{\textbf{Generalization ability test under the same simulation temporal length but different temporal resolution.}} Here we train the MLP-based SNNs and enhanced RNNs on N-MNIST under $dt=3ms$ and test them under $dt=\{3ms,~2ms,~1ms\}$. \label{fig:generalization} 
\end{figure}

Next, we do a simple experiment to test the generalization ability of models under different temporal resolutions. We train an SNN model, a vanilla RNN model (rate-coding-inspired), and an LSTM model (rate-coding-inspired) on N-MNIST under $dt=3ms$, and then test the trained models under $dt=\{3ms,~2ms,~1ms\}$. Also, we keep the simulation temporal length identical as above, fixing $T\times dt=45ms$ here. Unless otherwise specified, the leakage factor equals 0.3. Table \ref{tab:acc_generalization} reports the accuracy results, and the training curves can be found in Figure \ref{fig:generalization}. We have two observations: (1) the testing under $dt=2ms$ and $dt=1ms$ loses accuracy, and the degradation increases significantly as $dt$ becomes much smaller such as $dt=1ms$; (2) the SNN model presents better generalization ability. Specifically, when testing under $dt=1ms$, the SNN model only loses 2.18\% accuracy, while the vanilla RNN model and the LSTM model lose 19.82\% and 20.87\% accuracy, respectively, which are much higher than the loss of the SNN model.

We explain the above robustness of SNNs as follows. First, as mentioned earlier, SNNs are naturally suited for processing sparse features under smaller temporal resolution owing to the event-driven paradigm. Second, different from the trainable cross-neuron recurrent weights in RNNs, SNNs use self-neuron recurrence with restricted weights (i.e. the leakage factor $-e^{-\frac{dt}{\tau}}$). This recurrence restriction stabilizes the SNN model thus leading to improved generalization. To evidence the latter prediction, we additionally test the performance of the SNN model with trainable cross-neuron recurrent weights and present the results in Table \ref{tab:acc_generalization}. As expected, the generalization ability dramatically degrades, like RNNs. This might be caused by the increased number of parameters and more complex dynamics after introducing the trainable cross-neuron recurrence. Additionally, we try to identify whether the leakage factor would affect the generalization ability of SNNs. In all previous experiments, the leakage factor is fixed at 0.3; by contrast, we further test an SNN model with adaptive leakage factors by fixing only $\tau$ but varying $dt$. Also from Table \ref{tab:acc_generalization}, it can be seen that the adaptive leakage factor just slightly improves the robustness.

\subsection{Loss Function \& Temporal Contrast}

\begin{figure}[!htbp]
\centering     
\includegraphics[width=0.485\textwidth]{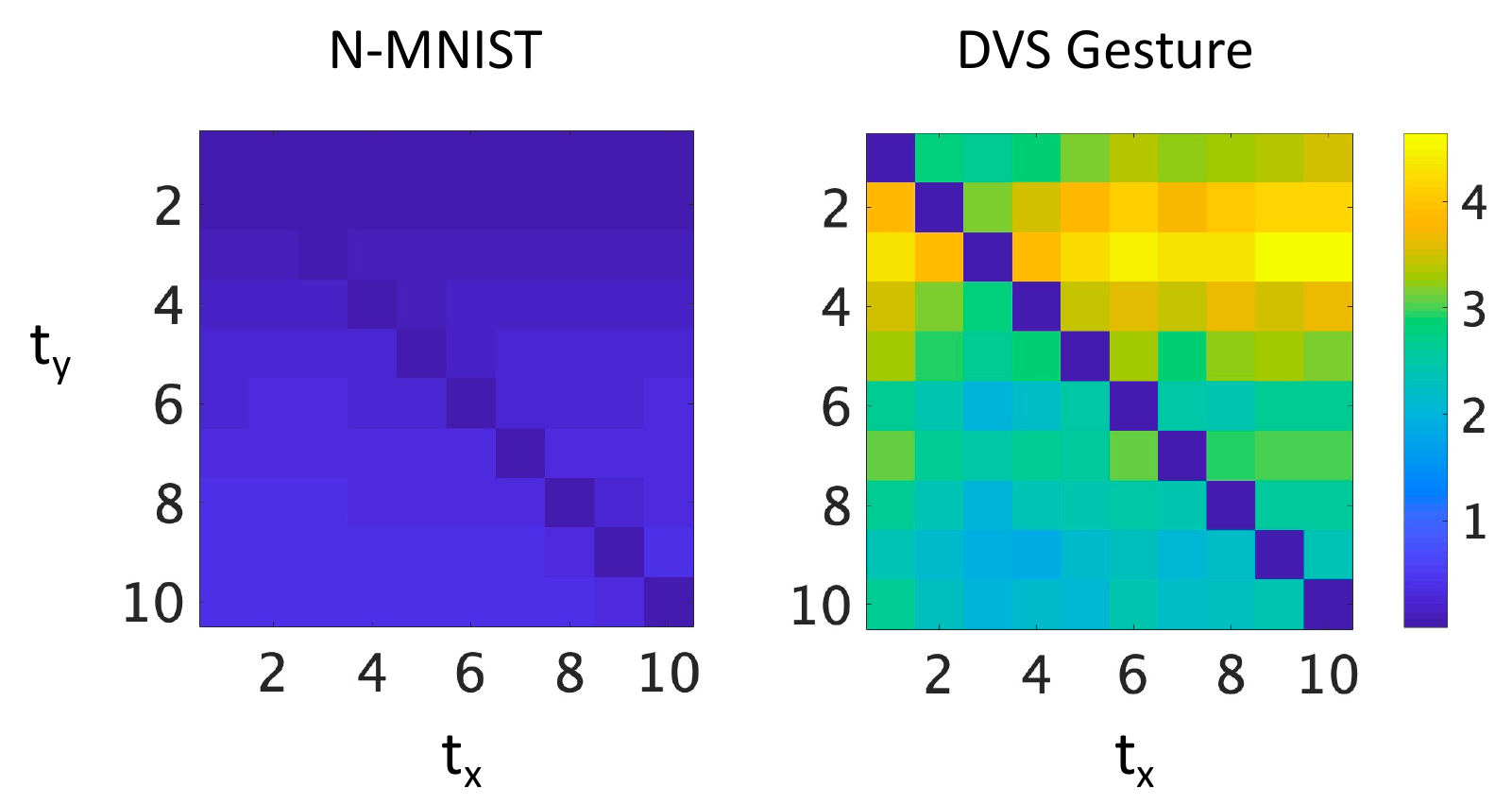}
\caption{\textbf{Visualization of the cross-entropy matrices calculated from N-MNIST and DVS Gesture.} The colorful block at the coordinate $(t_x,~t_y)$ represents the cross-entropy value between $\pmb{S}[t_x,t_x+k]$ and $\pmb{S}[t_y,t_y+k]$. Here we set $k=4$. } 
\label{fig:temporal_CE} 
\end{figure}

In Section \ref{sec:result_acc}, we have observed that the rate-coding-inspired loss function can boost more accuracy on DVS-captured datasets. In this subsection, we do a deeper analysis on this phenomenon. We define the temporal contrast of a neuromorphic vision dataset as the cross-entropy between slices at different timesteps. Specifically, we denote $\pmb{S}[t,t+k]$ as the slices between the $t$-th timestep  and the $(t+k)$-th timestep. Thus, there exists a cross-entropy value between $\pmb{S}[t_x,t_x+k]$ and $\pmb{S}[t_y,t_y+k]$ where $t_x$ and $t_y$ can be any two given timesteps. Here we define the cross-entropy value as
\begin{equation}
\label{equ:temporal_CE}
\begin{aligned}
CE(t_x,t_y) &= -\frac{1}{N}\sum_i(S_i[t_x,t_x+k] log^{\epsilon}(S_i[t_y,t_y+k])\\
&+ (1-S_i[t_x,t_x+k]) log^{\epsilon}(1-S_i[t_y,t_y+k]))
\end{aligned}
\end{equation}
where $i$ and $N$ are the index and number of elements in $\pmb{S}$, respectively. Note that $log^{\epsilon}$ is the $log$ function with numerical optimization: $log^{\epsilon}(x)=log(g^{\epsilon}(x))$. We set
\begin{equation}
\label{equ:CE_log}
\begin{cases}
g^{\epsilon}(x)=\epsilon,~&if~x\in [0,~\epsilon) \\
g^{\epsilon}(x)=x,~&if~x\in [\epsilon,~1-\epsilon] \\
g^{\epsilon}(x)=1-\epsilon,~&if~x\in (1-\epsilon,~1]
\end{cases}
\end{equation}
where $\epsilon=1e^{-16}$ is a small constant. The reason we do the above optimization is because the elements in $\pmb{S}$ can only be binary values within $\{0,~1\}$, which might cause zero or negative infinity results when passing through the $log$ function. Then, we visualize the cross-entropy matrices of the two neuromorphic vision datasets we use: the DVS-converted N-MNIST and the DVS-captured DVS Gesture, as presented in Figure \ref{fig:temporal_CE}.

\begin{table}[!htbp]
\caption{Mean and variance of the cross-entropy matrices.}
\label{tab:temporal_CE}
\vspace{2pt}
\centering
\renewcommand\arraystretch{1.3}
\resizebox{0.32\textwidth}{!}{
\begin{tabular}{ccc}
\hline\hline
  & \textbf{N-MNIST} & \textbf{DVS Gesture}   \\ \hline
\textbf{Mean}  & 0.2304 & 2.6912 \\
\textbf{Variance} & 0.0241 & 1.2701 \\ 
\hline\hline
\end{tabular}}
\end{table}

Apparently, it can be seen that the temporal contrast of DVS-captured datasets is much larger than that of DVS-converted datasets. This indicates that there are more temporal components in DVS-captured datasets, while the slices at different timesteps in DVS-converted datasets are close. Furthermore, we provide the statistic data, including mean and variance, of the cross-entropy matrices derived from the above two datasets. The calculation rules of mean and variance follow the normal definitions in statistics. As shown in Table \ref{tab:temporal_CE}, it also demonstrates that the data variance of DVS-captured datasets is much larger than that of DVS-converted datasets, which is consistent with the conclusion from Figure \ref{fig:temporal_CE}. By taking the outputs at different timesteps into account, the rate-coding-inspired loss function in Equation (\ref{equ:RNN_loss3}) is able to provide error feedback at all timesteps thus optimizing the final recognition performance. The above quantitative analysis can well explain the underlying reason that the rate-coding-inspired loss function can gain more accuracy boost on DVS Gesture than the gain on N-MNIST. We should note that, when the temporal contrast is too large, the effectiveness of the rate-coding-inspired loss function might be degraded due to the divergent gradient directions at different timesteps, which needs more practice in the future.

\subsection{Number of Parameters and Operations}

Besides the accuracy analysis, we further consider the memory and compute costs during model running. For the computational complexity, we take one layer with $M$ neurons as an example. In the forward pass, we count the operations when it propagates the activities to the next timestep and the next layer with $N$ neurons; while in the backward pass, we count the operations when it receives the gradients from the next timestep and the next layer. Notice that we only count the matrix operations because they occupy much more complexity than the vector and scalar ones. The comparison is presented in Table \ref{tab:compute_complexity}, which is mainly derived from Equation (\ref{equ:vanilla_RNN_layer})-(\ref{equ:LSTM_bp2}). Apparently, the SNN model consumes fewer operations owing to the self-neuron recurrence and avoids costly multiplications in the forward pass owing to the spike format. Furthermore, the event-driven computation in the forward pass can further reduce the required operations that are proportional to the normalized spike rate.

\begin{table}[!htbp]
\caption{Computational complexity comparison. ADDs -- additions, MULs -- multiplications, MACs -- multiplications and additions, $\alpha$ -- normalized spike rate.}
\label{tab:compute_complexity}
\vspace{2pt}
\centering
\renewcommand\arraystretch{1.3}
\resizebox{0.49\textwidth}{!}{
\begin{tabular}{cccc}
\hline \hline
\textbf{Data Path}   & \textbf{SNN}   & \textbf{Vanilla RNN} & \textbf{LSTM}  \\ \hline
Forward & $O(\alpha MN)$ ADDs & $O(MN+M^2)$ MACs & $O[4(MN+M^2)]$ MACs \\
Backward & $O(MN)$ MACs & $O(MN+M^2)$ MACs & $O(8MN)$ MULs + $O(MN+M^2)$ MACs \\
\hline\hline
\end{tabular}}
\end{table}

\begin{figure}[!htbp]
\centering     
\includegraphics[width=0.48\textwidth]{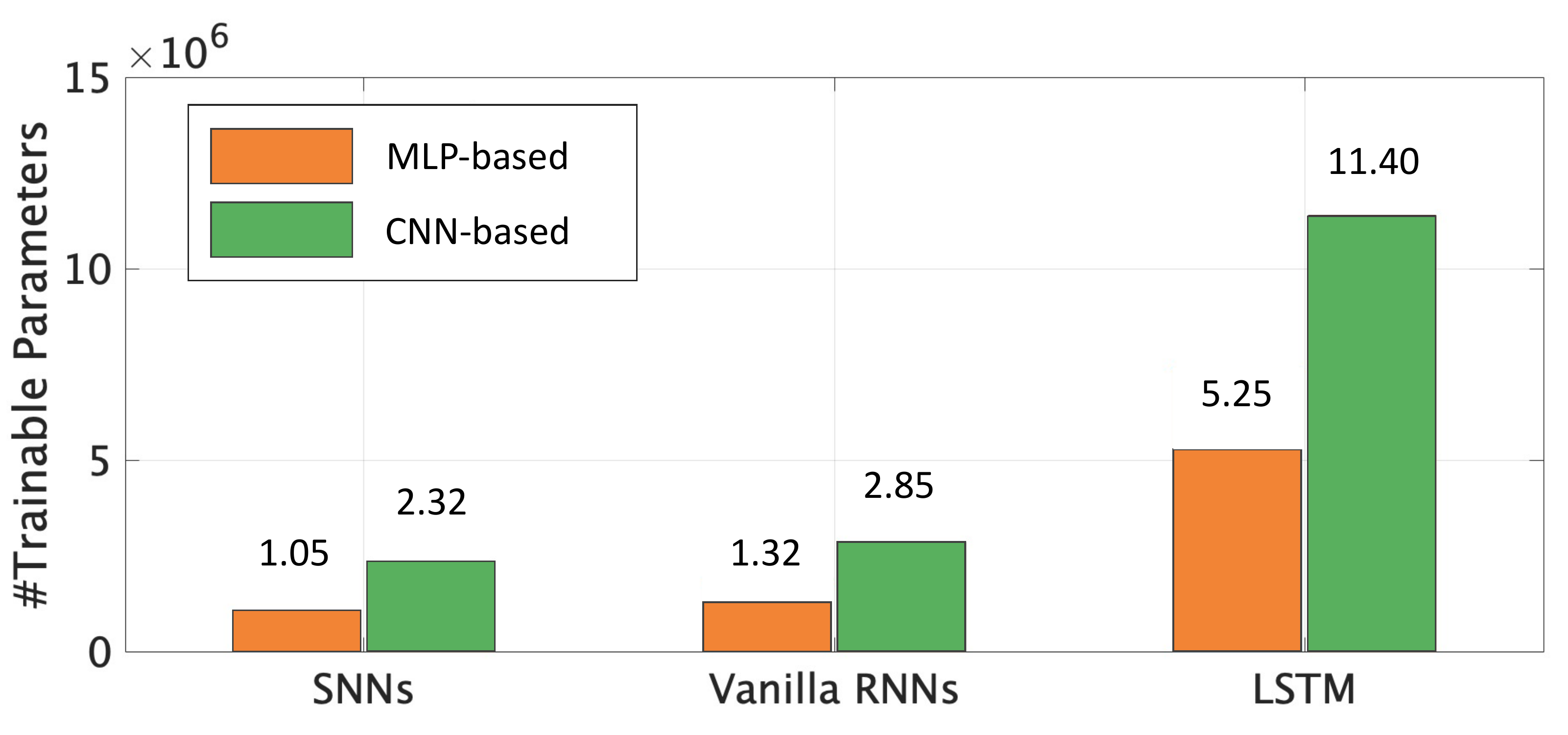}
\caption{\textbf{Number of model parameters on DVS Gesture.}} \label{fig:parameters} 
\end{figure}

On the other hand, the memory overhead is mainly determined by the number of parameters, especially when performing inference with only the forward pass on edge devices. Figure \ref{fig:parameters} compares the number of model parameters of SNNs, vanilla RNNs, and LSTM. Here we take the models we used on the DVS Gesture dataset for illustration. We find that the parameter amount of SNNs is much smaller than those of RNNs. Overall, SNNs only occupy about 80\% and 20\% parameters compared with the vanilla RNNs and LSTM, respectively.

Interestingly, despite the fewer operations and parameters of SNNs, the extra membrane potential path helps them achieve comparable (under large temporal resolution) or even better (under small temporal resolution) recognition accuracy than LSTM with complex gates; in the meantime, the self-neuron recurrence and the restricted recurrent weights make them more lightweight and robust.

\subsection{Discussion on Audio Data}

In the previous content, we focus on vision datasets. Actually, another important branch of data sources with spatiotemporal components is the audio data, which has also been used in neuromorphic computing \cite{wu2018spiking, wu2020deep}. To gently extend the scope of this work, we provide an extra experiment on an audio dataset in this subsection. 

\begin{table}[!htbp]
\caption{Accuracy results on audio data.}
\label{tab:acc_audio}
\vspace{2pt}
\centering
\renewcommand\arraystretch{1.3}
\resizebox{0.485\textwidth}{!}{
\begin{tabular}{cccc}
\hline \hline
 \multirow{2}*{$T$ \textbf{during Testing}}   & \multirow{2}*{\textbf{SNNs}} & \textbf{Vanilla RNNs} & \textbf{LSTM} \\ 
 & & \textbf{(Rate-Coding-inspired)} & \textbf{(Rate-Coding-inspired)} \\ \hline
 75 & 98.14\% & 96.81\% & 97.59\% \\ 
 25 & 89.23\% & 72.86\% & 87.55\% \\
 15 & 77.82\% & 46.14\% & 68.27\% \\
\hline\hline
\end{tabular}}
\end{table}

We select the Spoken-Digits \cite{Dua:2019} for testing. The network structure is ``Input-512FC-10''. The hyper-parameter setting is the same as that on N-MNIST except for the number of simulation timesteps, i.e. $T$. We set $T=75$ during training, while varying it during testing to explore the generalization. The results are listed in Table \ref{tab:acc_audio}. It can be seen that the vanilla RNNs perform the worst while SNNs are the best. Furthermore, SNNs show better generalization ability on this dataset, which is consistent with the observation in Section \ref{sec:generalization}.

%% file: text/Conclusion.tex
\section{Conclusion}\label{sec:conclusion}

In this work, we conduct a systematic investigation to compare SNNs and RNNs on neuromorphic vision datasets and then compare their performance and complexity. To make SNNs and RNNs comparable and improve fairness, we first identify several similarities and differences between them from the modeling and learning perspectives, and then unify the dataset selection, temporal resolution, learning algorithm, loss function, network structure, and training hyper-parameters. Especially, inspired by the rate coding scheme of SNNs, we modify the mainstream loss function of RNNs to approach that of SNNs; to test model robustness and generalization, we propose to tune the temporal resolution of neuromorphic vision datasets. Based on a series of contrast experiments on N-MNIST (a DVS-converted dataset) and DVS Gesture (a DVS-captured dataset), we achieve extensive insights in terms of recognition accuracy, feature extraction, temporal resolution and contrast, learning generalization, computational complexity and parameter volume. For better readability, we summarize our interesting findings as below:
\begin{itemize}
\item SNNs are usually able to achieve better accuracy than common RNNs. Whereas, the rate-coding-inspired loss function can boost the accuracy of RNNs especially LSTM to be comparable or even slightly better than that of SNNs.

\item The event-driven paradigm of SNNs makes them more suitable to process sparse features. Therefore, in the cases of small temporal resolution with sparse spike events, SNNs hold obvious accuracy superiority.

\item On one hand, LSTM can memorize long-term dependence via the complex gates, while the extra membrane potential path of SNNs also brings longer-term memory than vanilla RNNs; on the other hand, the temporal contrast of slices in DVS-captured datasets is much larger than that in DVS-converted datasets, thus the processing of DVS-captured datasets depends more on the long-term memorization ability. These two sides can explain the reason that SNNs and LSTM significantly outperform vanilla RNNs on DVS Gesture, while this gap is small on N-MNIST.

\item The self-neuron recurrence pattern and restricted recurrent weights of SNNs greatly reduce the number of parameters and operations, which improves both the running efficiency and the model generalization. 
\end{itemize}

We believe that the above conclusions can benefit the neural network selection and design on different workloads in the future. We simply discuss several examples. On DVS-converted datasets, the accuracy gap between different models is small so that any model selection is acceptable. On DVS-captured datasets, we do not recommend vanilla RNNs due to the low accuracy. When the temporal resolution is large, we recommend LSTM with the rate-coding-inspired loss function; while when the temporal resolution is small, we recommend SNNs. If we need a compact model size, we always recommend SNNs that have significantly fewer parameters and operations. Moreover, it might be possible to improve models by borrowing ideas from each other. For instance, vanilla RNNs can be further enhanced by introducing more information propagation paths like the membrane potential path in SNNs; LSTM can be made more compact and robust by introducing the recurrence restriction; SNNs can be improved by introducing more gates like LSTM. It is even possible to build a hybrid neural network model by combining multiple kinds of neurons, thus taking the advantages of different models and alleviating their respective defects. In addition, we mainly focus on vision datasets and just provide very limited exploration on audio data in this work. More extensive experiments in a wide spectrum of tasks are highly expected.